\documentclass{article}
\pdfoutput=1

\usepackage{arxiv}

\usepackage[utf8]{inputenc} % allow utf-8 input
\usepackage[T1]{fontenc}    % use 8-bit T1 fonts 
\usepackage{booktabs}       % professional-quality tables
\usepackage{amsfonts}       % blackboard math symbols
\usepackage{nicefrac}       % compact symbols for 1/2, etc.
\usepackage{microtype}      % microtypography
\usepackage{lipsum}
\usepackage{amsthm}
\usepackage{booktabs}  
\usepackage{siunitx}
\usepackage{makeidx}
\usepackage{blindtext}
\usepackage{array} 
\usepackage{amssymb}
\usepackage{enumerate} 
\usepackage{subcaption} 
\usepackage{amssymb}
\usepackage{amsopn}
\usepackage{algorithm}
\usepackage{algorithmicx}
\usepackage{algpseudocode}
\usepackage{bm} 
\usepackage{rotating}
\usepackage{multicol}
\usepackage{multirow} 
\usepackage{tabularx}

\usepackage{graphicx}
\usepackage{epstopdf}
\usepackage{hyperref}

\usepackage{comment}

\usepackage{arydshln}
\usepackage{dsfont} 
\usepackage{color} 

\usepackage[misc,geometry]{ifsym}

\definecolor{DSgray}{cmyk}{0,1,0,0}

 \newtheorem{definition}{Definition}
 
 \newtheorem{lemma}{Lemma}
 \newtheorem{corollary}{Corollary}
 
\newtheorem{theorem}{Theorem}
\newtheorem{observation}[theorem]{Observation}

\DeclareMathOperator*{\argmin}{argmin}

\title{Improving the Effectiveness and Efficiency of Stochastic Neighbour Embedding with Isolation Kernel}

\author{
  Ye Zhu$^{ (\textrm{\Letter})}$%\thanks{This work has been submitted to the IEEE for possible publication. Copyright may be transferred without notice, after which this version may no longer be accessible.} 
  \\
  School of Information Technology\\
  Deakin University\\
  Victoria, Australia 3125 \\
  \texttt{ye.zhu@ieee.org} \\ 
   \And
  Kai Ming Ting\\
  National Key Laboratory for Novel Software Technology\\
  Nanjing University\\
  Jiangsu, China 210023 \\
  \texttt{tingkm@nju.edu.cn} \\  
}

\begin{document}
\maketitle

\begin{abstract}

This paper presents a new insight into improving the performance of Stochastic Neighbour Embedding (t-SNE) by using Isolation kernel instead of Gaussian kernel. Isolation kernel outperforms Gaussian kernel in two aspects. First, the use of Isolation kernel in t-SNE overcomes the drawback of misrepresenting some structures in the data, which often occurs when Gaussian kernel is applied in t-SNE. This is because Gaussian kernel determines each local bandwidth based on one local point only, while Isolation kernel is derived directly from the data based on space partitioning. Second, the use of Isolation kernel yields a more efficient similarity computation because {\em data-dependent} Isolation kernel has only one parameter that needs to be tuned. In contrast, the use of {\em data-independent}  Gaussian kernel increases the  computational cost by determining $n$ bandwidths for a dataset of $n$ points. As the root cause of these deficiencies in t-SNE is Gaussian kernel, we show that simply replacing Gaussian kernel with Isolation kernel in t-SNE significantly improves the quality of the final visualisation output (without creating misrepresented structures) and removes one key obstacle that prevents t-SNE from processing large datasets. Moreover, Isolation kernel enables t-SNE to deal with large-scale datasets in less runtime without trading off accuracy, unlike existing methods in speeding up t-SNE.
\end{abstract}

% keywords can be removed

\section{Introduction and Motivation}
\label{sec_intro}
t-SNE \cite{TSNE} has been a successful and popular dimensionality reduction method for visualisation. It aims to project high-dimensional datasets into lower-dimensional spaces while preserving the similarities between data points, as measured by the KL divergence. The original SNE \cite{SNE} employs a Gaussian kernel to measure similarity in both high and low-dimensional spaces. t-SNE replaces the Gaussian kernel with the distance-based similarity $(1+d_{ij})^2$ (where $d_{ij}$ is the distance between instances $i$ and $j$) in low-dimensional space, while retaining the Gaussian kernel for  high-dimensional space. % The distance-based similarity has a heavy-tailed distribution that alleviates issues related to far points and optimisation in SNE \cite{TSNE}.

When using the Gaussian kernel, t-SNE has to fine-tune a bandwidth of the Gaussian kernel centred at each point in the given dataset because Gaussian kernel is independent of data distribution. In other words, t-SNE must determine $n$ bandwidths for a dataset of $n$ points. 

If we look into the bandwidth determination process, it is accompanied by using a heuristic search with a single global parameter called perplexity such that the Shannon entropy is fixed for all probability distributions at all points in adapting each bandwidth to the local density of the dataset. As the perplexity can be interpreted as a smooth measure of the effective number of neighbours \cite{TSNE}, the method can be interpreted as using a user-specified number of nearest neighbours (aka kNN) in order to determine the $n$ bandwidths (more on this point in the discussion section.) Whilst there is a single external parameter $perplexity$,  a bandwidth setting  must be optimised for each data point internally.

This becomes the first obstacle in dealing with large datasets due to massive computational cost of the bandwidth search process. In addition, the point-based bandwidth is also the cause of misrepresentation in high-dimensional space under some conditions. 

To date, the common practice is still using Gaussian kernel in t-SNE on  high-dimensional datasets. However, sound and workable solutions to its drawbacks mentioned above have not been brought up yet. The contributions of this paper are: 

\begin{enumerate}[(1)]
    \item Uncovering two deficiencies due to the use of the Gaussian kernel. First, the point-based-bandwidth Gaussian kernel often creates misrepresented structure(s) which do not exist in high-dimensional space under some conditions.
    Second, the use of the data-independent kernel requires t-SNE to determine $n$ bandwidths for a dataset of $n$ points, despite the fact that a user needs to set one parameter only. This becomes one key obstacle in dealing with large datasets.
    \item Revealing the advantages of using a partition-based data-dependent kernel in t-SNE. First, this kernel represents the true structure(s) in the high-dimensional space under the same condition mentioned above. Second, the data-dependent similarity is set with a single parameter only; this allows it to be computed more efficiently. This enables t-SNE to deal with large-scale datasets without trading off accuracy with faster runtime, without resorting to approximation methods.
    \item Proposing an improvement to t-SNE by simply replacing the data-independent kernel with a data-dependent kernel, leaving the rest of the procedure unchanged.
    \item Verifying the effectiveness and efficiency of the data-dependent kernel in t-SNE. 
\end{enumerate}

The adopted data-dependent kernel is Isolation kernel \cite{ting2018IsolationKernel,aNNE} and the experiment result shows that using Isolation kernel will improve the performance of t-SNE and solve the issues brought by Gaussian kernel in t-SNE.

The rest of the paper is organised as follows.  The current t-SNE and related work are described in Section~\ref{sec_t-SNE}. The deficiencies of using Gaussian kernel is presented in Section \ref{sec_deficencies}. In Section~\ref{sec_IK1}, we characterise the selected Isolation kernel and Section~\ref{sec_evaluation} presents the empirical evaluation of using Isolation kernel in t-SNE. Discussion and conclusions are given in the last two sections.

\section{Background}
\label{sec_t-SNE}
\subsection{Basics  of t-SNE }

Given a dataset $D=\{x_1, \dots, x_n \}$ in $\mathbb{R}^d$, t-SNE aims to map $D \in \mathbb{R}^d$ to $D' \in \mathbb{R}^{d'}$ where $d' \ll d$ such that the similarities between points are preserved as much as possible from the high-dimensional space to the low-dimensional space. As t-SNE is meant for a visualisation tool, $d'=2$ usually.

The similarity between a pair of points $x_i$, $x_j$ (resp. $x'_i$ $x'_j$) in a high (resp. low)-dimensional space is measured by a probability $p_{ij}$ (resp. $p'_{ij}$) that point $x_i$ picks $x_j$ as its neighbour. The probability distributions are computed based on distance measures between the points in the respective space. The aim of this family of projection methods is to project the points from $x$ to $x'$ in such a way that the probability distributions between $p_{ij}$ and $p'_{ij}$ are as similar as possible.

The similarity between $x_i$ and $x_j$ is measured using a Gaussian kernel as follows:
\begin{equation}
\mathcal{K}(x_i,x_j) = exp(\frac{-\parallel x_i - x_j \parallel^2}{2\sigma_i^2}) 
\end{equation}

t-SNE computes the conditional probability $p_{j|i}$ that $x_i$ would pick $x_j$ as its neighbour as follows:
\begin{equation}
p_{j|i} = \frac{\mathcal{K}(x_i,x_j)}{\sum_{k \ne i} \mathcal{K}(x_i,x_k)} 
\label{eqn_p}
\end{equation}

The probability $p_{ij}$, a symmetric version of $p_{j|i}$, is computed as:
\begin{equation}
p_{ij} = \frac{p_{j|i} + p_{i|j}}{2n} 
\label{eqn_p2}
\end{equation}

t-SNE performs a binary search for the best value of $\sigma_i$ such that the perplexity of the conditional distribution equals a fixed perplexity specified by the user. Therefore, the bandwidth is adapted to the density of the data, i.e., small (large) values of $\sigma_i$ are used in dense (sparse) regions. The perplexity is defined as:

\begin{equation}
    Perp(P_i)=2^{H(P_i)} 
\end{equation}
\noindent
where $P_i$ represents the conditional probability distribution over all other data points given data point $x_i$ and $H(P_i)$ is the Shannon entropy:

\begin{eqnarray} 
    H(P_i)=-\sum_{j}p_{j|i} \ log_{2} \ p_{j|i} 
\end{eqnarray}

The perplexity is a smooth measure of the effective number of neighbours, similar to the number of nearest neighbours $k$ used in $k$NN methods \cite{SNE}. Thus, $\sigma_i$ is adapted to the density of the data, i.e., it becomes small for dense data since the $k$-nearest neighbourhood is small and vice versa. In addition, \cite{TSNE} point out that there is a monotonically increasing relationship between perplexity and the bandwidth $\sigma_i$.

The similarity between $x'_i$ and $x'_j$ in the low-dimensional space is measured as:
\[ s(x'_i,x'_k) = (1 + \parallel x'_i - x'_j \parallel^2)^{-1} \]  

and the corresponding probability is defined as:
\[ p'_{ij} = \frac{s(x'_i,x'_j)}{\sum_{k \ne \ell} s(x'_\ell,x'_k)} \]

The distance-based similarity $s$ is used because it has heavy-tailed distribution, i.e., it approaches an inverse square law for large pairwise distances. This means that the far apart mapped points have $p'_{ij}$ which are almost invariant to changes in the scale of the low-dimensional space \cite{TSNE}.
 
Note that the probability distributions are defined in such a way that $p_{ii} =0$ and $p'_{ii}=0$,  i.e. a node does not pick itself as a neighbour. 

The location of each point $x' \in D'$ is determined by minimising a cost function based on the (non-symmetric) Kullback-Leibler divergence of the joint probability distribution $P'$  in the low-dimensional space from the joint distribution $P$ in the high-dimensional space:

\[ KL(P \parallel P') =  \sum_{i \ne j} p_{ij} \log \frac{p_{ij}}{p'_{ij}} \]

The use of the Gaussian kernel $\mathcal{K}$ sharpens the cost function in retaining the local structure of the data when mapping from the high-dimensional space to the low-dimensional space. The main computational step in applying t-SNE is to determine the value of bandwidth $\sigma$ for each data point.

The procedure of t-SNE is provided in Algorithm \ref{alg_t-SNE}. Note that $m=n$ for small datasets. For large datasets, $m \ll n$; and this is to be discussed in Section \ref{sec_large_datasets}.

    \begin{algorithm}[!htbp]
        \caption{t-SNE$(D, Perp, m)$}
        \begin{algorithmic}[1]  
            \Require $D$ - Dataset $\{x_1,\dots,x_n \}$; $Perp$ - Perplexity
        \State Determine $\sigma_i$ for every $x_i \in D$ based on $Perp$
        \State Compute matrix $[p_{ij}]_{m \times m}$ according to Equations \ref{eqn_p} \& \ref{eqn_p2}
%        \State Compute $p_{j|i}$ based on Gaussian kernel $\mathcal{K}(x_i,x_j)$
%        \State Set $p_{ij} = \frac{p_{j|i} + p_{i|j}}{2n}$
        
        \State Compute low-dimensional $D'$ and $p'_{ij}$ which minimise the KL divergence
        \State Output low-dimensional data representation $D'=\{x'_1, \dots, x'_m\}$
        \end{algorithmic}
        \label{alg_t-SNE}
    \end{algorithm}
    
\subsection{Related work}

\label{sec_related_work}
t-SNE \cite{TSNE} and its variations have been widely applied in dimensionality reduction and visualisation. In addition to t-SNE \cite{TSNE}, which is one of the commonly used visualisation methods, many other variations have been proposed to improve SNE in different aspects.

There are improvements based on some revised Gaussian kernel functions in order to get better similarity measurements. \cite{cook2007visualizing} propose a symmetrised SNE; \cite{yang2009heavy} enable t-SNE to accommodate various heavy-tailed embedding similarity functions; and \cite{van2012stochastic} propose an algorithm based on similarity triplets of the form ``A is more similar to B than to C'' so that it can model the local structure of the data more effectively.

Based on the concept of information retrieval, NeRV \cite{venna2010information} uses a cost function to find a trade-off between precision and recall of ``making true similarities visible and avoiding false similarities'', when projecting data into 2-dimensional space for visualising similarity relationships. Unlike SNE which relies on a single Kullback-Leibler divergence, NeRV uses a weighted mixture of two dual Kullback-Leibler divergences in neighbourhood retrieval. Furthermore, JSE \cite{lee2013typeRNX} enables t-SNE to use a different mixture of Kullback-Leibler divergences, a kind of generalised Jensen-Shannon divergence, to improve the embedding result.

To reduce the runtime of t-SNE, \cite{van2014accelerating} explores tree-based indexing schemes and uses the Barnes-Hut approximation to reduce the time complexity to $\mathcal{O}(n log(n))$, where $n$ is the data size. This gives a trade-off between speed and mapping quality. To further reduce the time complexity to $\mathcal{O}(n)$, \cite{linderman2019fast} utilise a fast Fourier transform to dramatically reduce the time of computing the gradient during each iteration. The method uses vantage-point trees and approximates nearest neighbours in dissimilarity calculation with rigorous bounds on the approximation error. 

Some works focus on analysing the heuristics methods for solving non-convex optimisation problems for the embedding \cite{linderman2017clustering,shaham2017stochastic}. Recently, \cite{arora2018analysis} theoretically analyse this optimisation  and provide a framework to make clusterable data visually identifiable in the 2-dimensional embedding space. These works focus on changing the optimisation problem and are not related to similarity measurements. %They are not directly relevant to the work reported here.

So far, however, none of these studies has investigated the suitability of Gaussian kernel in t-SNE. The following two sections will uncover the issues of using Gaussian kernel in t-SNE and propose to replace it with Isoaltion kernel.

\section{Deficiencies of Gaussian kernel when used in t-SNE} 
\label{sec_deficencies}

Here we list two identified deficiencies of Gaussian kernel that cause poor visualisation outputs and high computational cost in t-SNE.

\subsection{The first deficiency}
\subsubsection{Point-based bandwidth: the cause of misrepresentation in high-dimensional space}
\label{sec_pint-based_Bandwidth}

As bandwidth $\sigma_i$ of the Gaussian kernel is fixed for each point $x_i$, we identify the following observation: 

\begin{observation}
Gaussian kernel with point-based bandwidth can misrepresent the structure of a data distribution, having points significantly denser than the majority of the points in a sample generated from the distribution. 
\end{observation}

Intuitively, as each point-based bandwidth represents one local density only, the Gaussian kernel can misrepresent the relationship between multiple clusters in the joint distribution of the overlap region. We provide two example cases in which misrepresentation occurs, i.e., there are multiple subspace clusters; each is a Gaussian distribution of the same mean with: (i) different variances; and (ii) the same variance.

Let $\mathcal{X}_1$ and $\mathcal{X}_2$ be two subspace regions in a high-dimensional space, and points in the two clusters are generated from the Gaussian distributions $N[0, v_1]$ and $N[0, v_2]$, respectively; and the distributions only overlap at the origin $O$. 

In case (i) where variance $v_1 \ll v_2$. Let point $x_{k1} \in \mathcal{X}_1$ be the point closest to $O$ in the dense cluster, and point $x_{k2} \in \mathcal{X}_2$ be the point closest to $O$ in the sparse cluster. Then, $\mathcal{K}(O, x_{k1}) \gg \mathcal{K}(O, x_{k2})$ because 
$\parallel O - x_{k1} \parallel \ll \parallel O - x_{k2} \parallel$ and $\mathcal{K}(O, \cdot)$ is inversely proportional to distance.

In case (ii) where $v_1 = v_2$,
 using an appropriate setting in the current t-SNE procedure, each point $x$ in either  $\mathcal{X}_1$ or $\mathcal{X}_2$ would have learned approximately the same bandwidth $\sigma$, except the origin $O$ because $O$ has at least double the density than any point in either cluster. As a result,  
$\forall x_i, x_j  \in \mathcal{X}_1$ (or $\forall x_i, x_j  \in \mathcal{X}_2$) and $\parallel O - x_i \parallel = \parallel x_j - x_i \parallel$, $\mathcal{K}(O, x_i) \ll \mathcal{K}(x_j, x_i)$ because $\sigma_O \ll \sigma_j$. This means that the origin is very dissimilar to any points in either cluster. 

Simulations of the two cases are given below:
\begin{enumerate}[(i)]
  \item  Five subspace clusters having different variances in a 50-dimensional space (see the simulation details in the footnote\footnote{The synthetic 50-dimensional dataset contains 5 subspace clusters. Each cluster has 250 points, sampled from a 10-dimensional Gaussian distribution with the other 40 irrelevant attributes having zero values; but these $4 \times 10$ attributes are relevant to the other four Gaussian distributions. In other words, no clusters share a single relevant attribute. In addition, all clusters have significantly different variances (the variance of the 5th cluster is 625 times larger than that of the 1st cluster). The first three clusters share the same mean; but the last two have different means. The five clusters have distributions: $N[0, 1]$, $N[0, 16]$, $N[0, 81]$, $N[400, 256]$ and $N[500, 625]$ in each dimension.}.)

Using Gaussian kernel, SNE creates a misrepresentation of the structure in the high-dimensional space. The simulation result is shown in
the first row in Table \ref{v0}: t-SNE is unable to identify the joint component of the three clusters in different subspaces which share the same mean at the origin only in the high-dimensional space but nowhere else. Notice that the mapped origin point is misrepresented to be associated with one cluster only; and it is totally disassociated with the other two clusters.

In contrast, the same t-SNE algorithm employing the Isolation kernel \cite{ting2018IsolationKernel,aNNE}, instead of a Gaussian kernel, produces the mapping which truly represents the structure in the high-dimensional space: the three clusters are well separated and yet they share some common points, indicated by the mapped origin point as shown in the second row in Table \ref{v0}.

  \begin{table} [!tbp] 
\setlength{\tabcolsep}{0pt}
	\centering
	\caption{Visualisation results of the t-SNE using Gaussian kernel and Isolation kernel  on a 50-dimensional dataset with 5 subspace clusters, each in a different 10-dimensional subspace. The black cross indicates the mapped point of the origin in the high-dimensional space shared by three clusters in different subspaces. Note that in (c), all points of the red cluster (cluster 1) are concentrated and they overlap  with the mapped origin. $perplexity$ and $\psi$ are the key parameters for Gaussian kernel and Isolation kernel, respectively.} 
	\begin{tabular}{cccc}
		\hline 
       \begin{turn}{90} \ \ Gaussian kernel \end{turn}&      \includegraphics[width=1.6in]{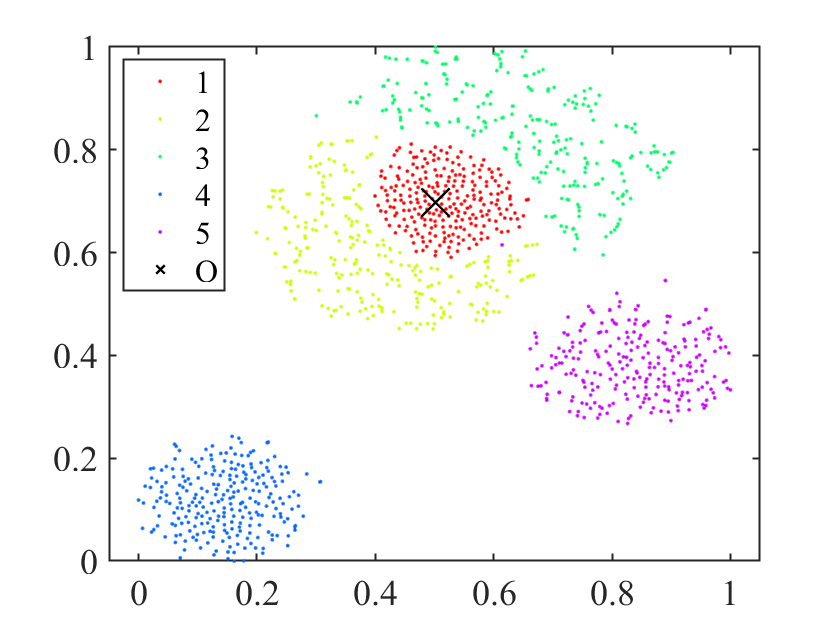}&
      \includegraphics[width=1.6in]{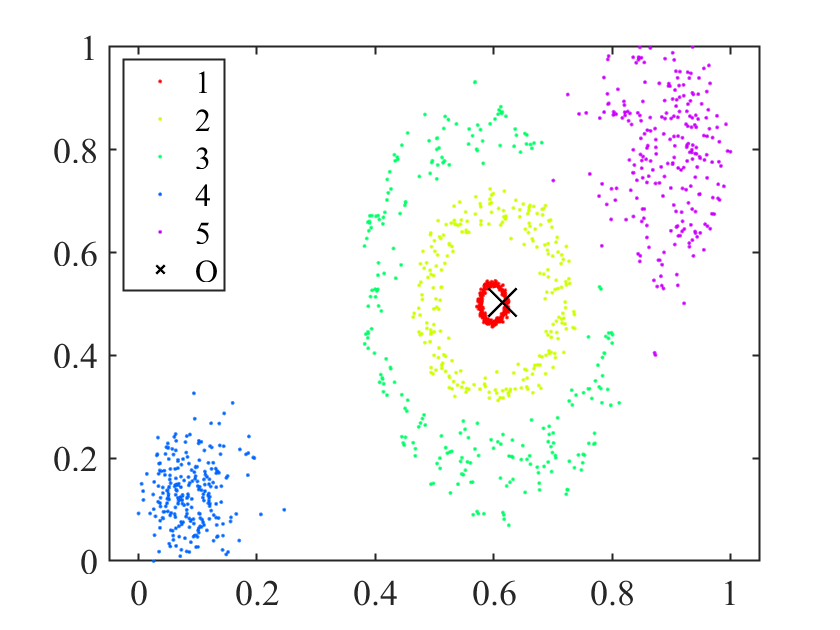} &
      \includegraphics[width=1.6in]{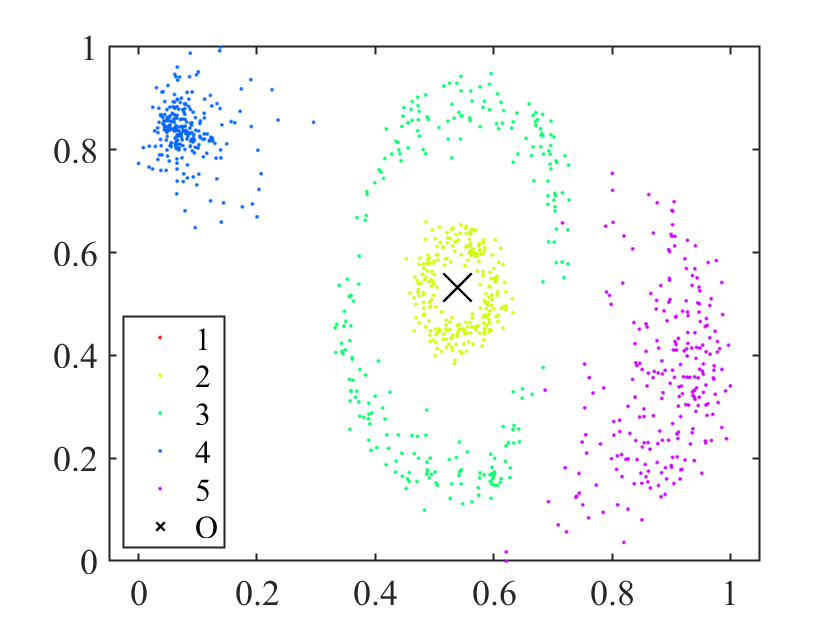}  \\     
       & (a) $perplexity=50$   & (b) $perplexity=250$ & (c) $perplexity=500$\\  \hdashline
       \begin{turn}{90} \quad  Isolation kernel \end{turn}&      \includegraphics[width=1.6in]{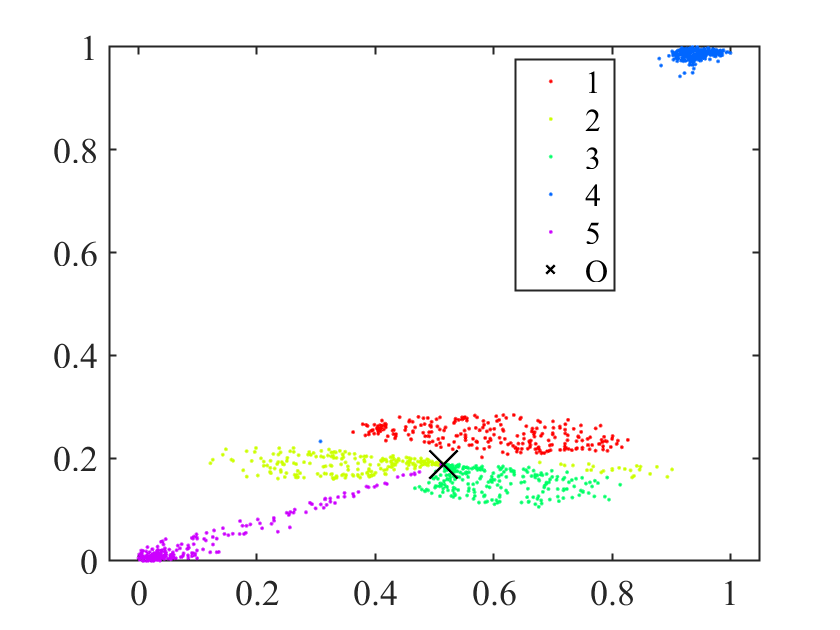} &
      \includegraphics[width=1.6in]{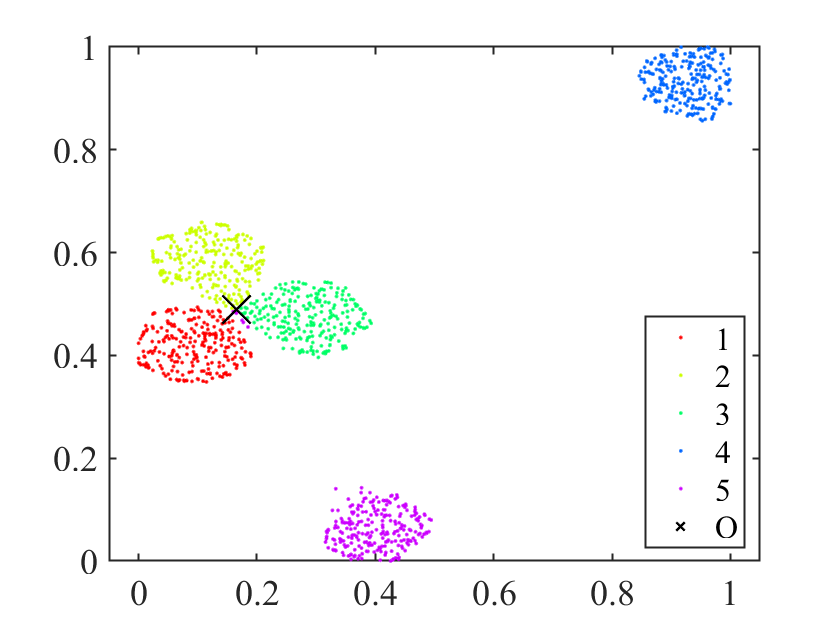} &
      \includegraphics[width=1.6in]{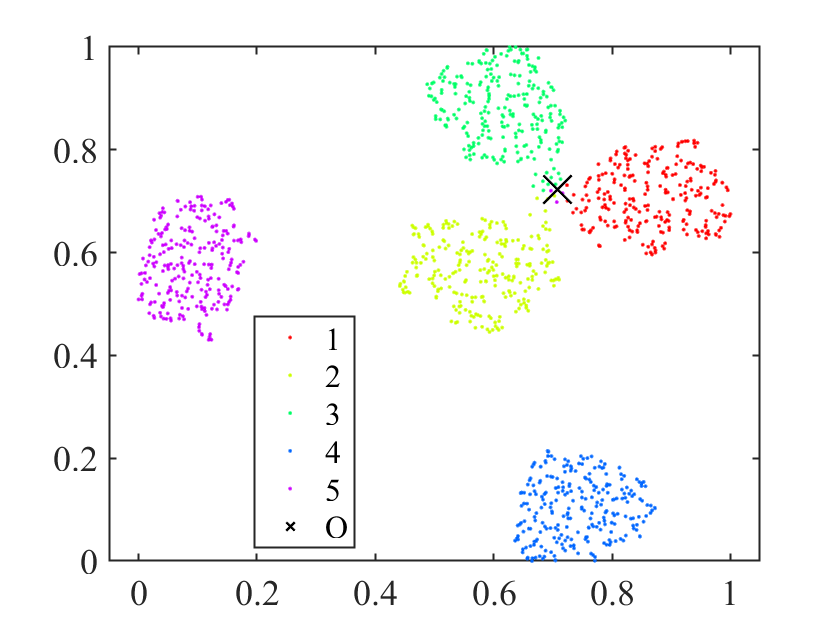}  \\    
      & (d) $\psi=50$ &  (e) $\psi=250$ & (f) $\psi=500$ \\
		\hline 
	\end{tabular}
	\label{v0}
\end{table}
 
\begin{table} [!htbp] 
%\vspace{10mm}
 % \renewcommand{\arraystretch}{1.2}
\setlength{\tabcolsep}{0pt}
\centering
\caption{Visualisation results of t-SNE with Gaussian kernel and Isolation kernel on a 200-dimensional dataset with two equal density subspace clusters. Note that in (c), the origin is far away from both clusters, although there is a clear gap between the two clusters. The green box in (c) presents a zoom-in view of the two clusters.} 
  \begin{tabular}{cccc}
    \hline 
       \begin{turn}{90} \ \ Gaussian kernel \end{turn}&      \includegraphics[width=1.6in]{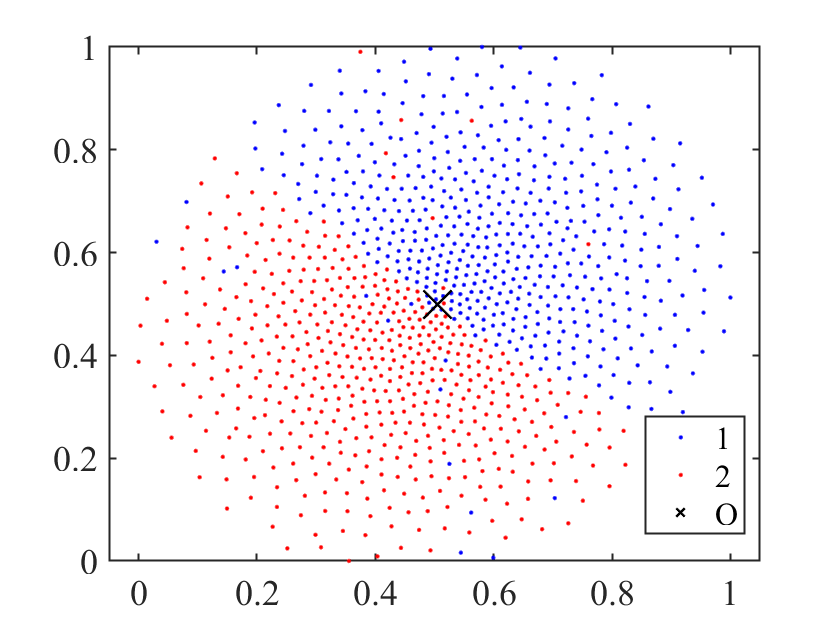}&
       % 0.56119      0.98005       2548.8
      \includegraphics[width=1.6in]{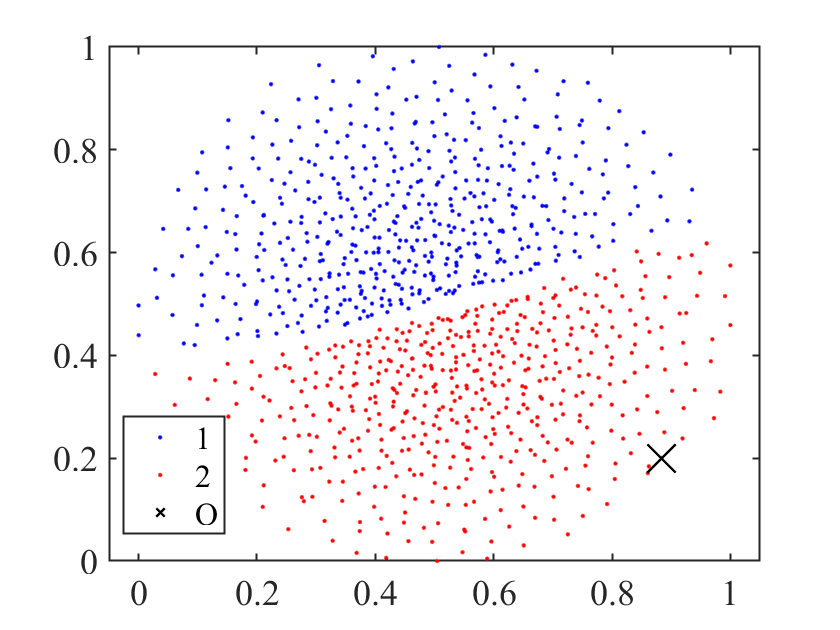} &
      % 0.17793       8.5203         1948
      \includegraphics[width=1.6in]{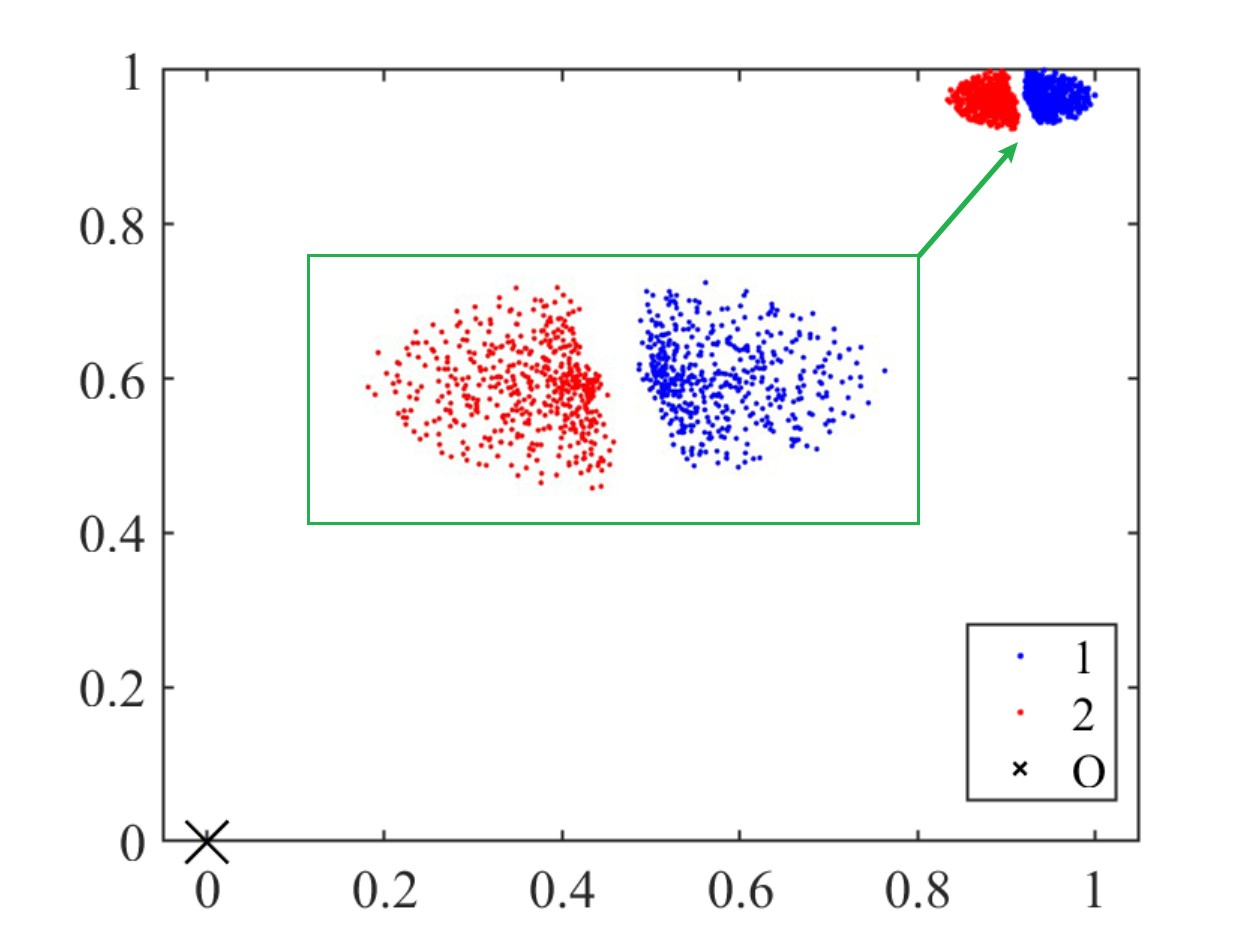}  \\     
     %      0.088882       7.7477       1923.2
       & (a) $perplexity=50$   & (b) $perplexity=210$ & (c) $perplexity=300$\\  \hdashline
       \begin{turn}{90} \quad  Isolation kernel \end{turn}&      \includegraphics[width=1.6in]{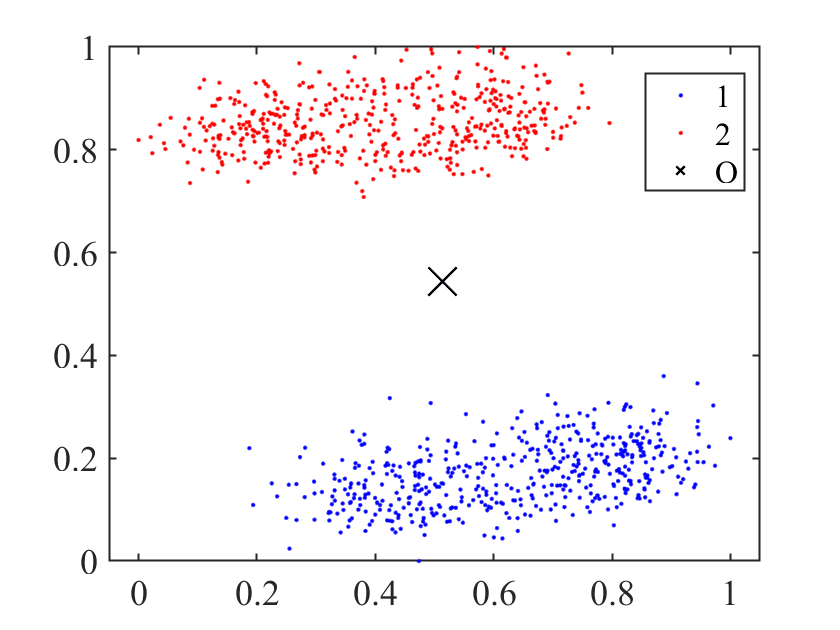} &
       %   0.063807      0.62081        16682
      \includegraphics[width=1.6in]{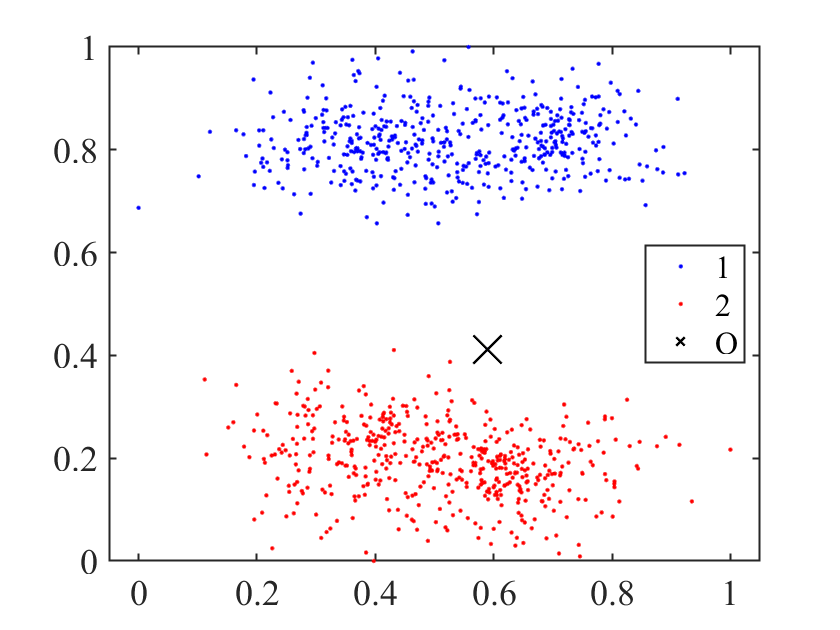} &
      % 0.21948      0.46932        10986
      \includegraphics[width=1.6in]{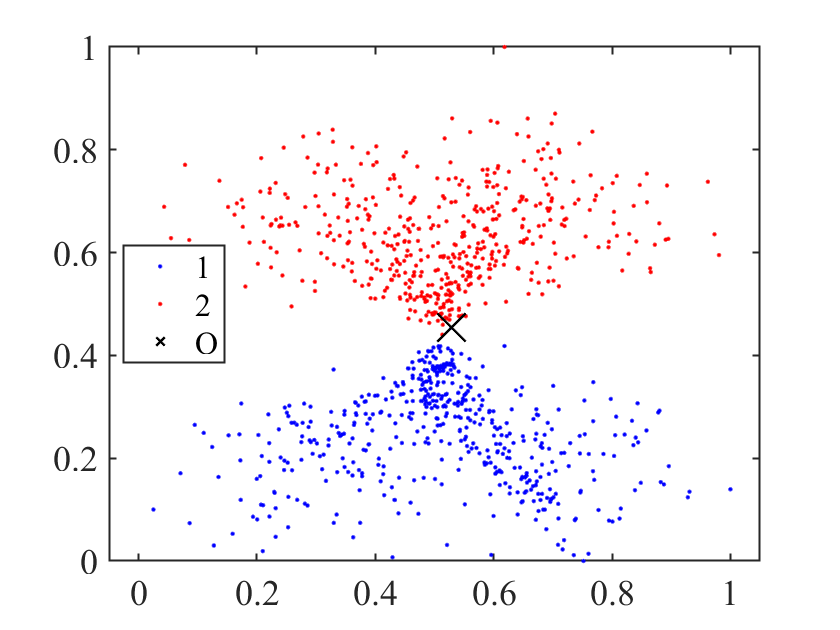}  \\    
      & (d) $\psi=50$ &  (e) $\psi=210$ & (f) $\psi=300$ \\
      %  0.52777      0.49528       4644.6
      \hline 
  \end{tabular}
\label{v03}
\end{table}

 \item Two subspace clusters in a 200-dimensional dataset with two subspace clusters having the same Gaussian distribution $N[0, 1]$ but in different subspaces\footnote{Each cluster has 500 points, sampled from a 100-dimensional Gaussian distribution $N[0, 1]$ with the other 100 irrelevant attributes having zero values; and no clusters share a single relevant attribute.}.

 Table \ref{v03} shows the simulation results.  When Gaussian kernel is used, the t-SNE with a small $perplexity$ produces small bandwidth for every point---leading to each point has almost the same low similarity with every other point in the dataset, as shown in Figure (a) in Table \ref{v03}. Note that the two clusters could not be distinguished in the visualization if the colors, indicating the ground truth labels, are not used in the plot. 
 Yet, the t-SNE with a large $perplexity$ produces large bandwidths for all points, except the origin which has a significantly smaller bandwidth---note that the origin (denoted as $\times$) and the rest of the points are at the opposite corners in Figure (c)  in Table \ref{v03}.  This is because the origin, being the only overlap point between the two clusters, has a significantly higher density than all other points. As both clusters have the same variance, all their points have low density (relative to the origin) are `learned' to have approximately the same bandwidth---which is significantly larger than that of the origin. As a result, the origin is very dissimilar to all other points; though all the other points are correctly clustered into two separate groups. Figure (b) shows the movement of the origin using a $perplexity$ between those used in Figure (a) \& Figure (c), i.e., the origin moves from in-between the two clusters in (a) to the edge of a cluster in (b); before moving to a location far away from both clusters in (c).    
 
 In contrast, when the Isolation kernel is used, the origin is always positioned in-between the two clusters, independent of the $\psi$ parameter setting. 

\end{enumerate}

Note the above-mentioned deficiency is not restricted to subspace clusters without shared attributes. An example using subspace clusters with shared attributes
can be found in Appendix A.

%%%%%%%%%%%%%%%%%%%%%%%%

%%%%%%%%%%%%%%%%%%%%%%%%%% 

\subsubsection{No need for point-based bandwidth in Isolation kernel}
The space partitioning mechanism of the Isolation kernel \cite{ting2018IsolationKernel,aNNE} determines the size of the partitions in the local region: it produces large partitions in the sparse region and small partitions in the dense region (see Section~\ref{sec_IK-vs-GK} for more details.) As it is partition-based, points in the local neighbourhood are most likely to be in the same partition. As such, points in the intersection of clusters (in different subspaces as shown in Table \ref{v0}) are almost always captured by the same partition of Isolation kernel.

\begin{figure}[h]
\centering
  \begin{subfigure}{.49\textwidth}
  \centering
    \includegraphics[width=1.5in]{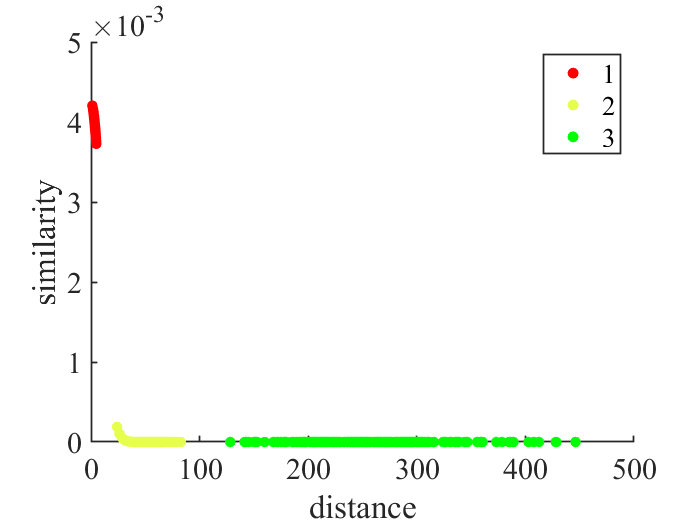}
   \caption{Gaussian kernel with $perplexity=250$}  
     \label{2a}
  \end{subfigure}  %
  \begin{subfigure}{.49\textwidth}
  \centering
    \includegraphics[width=1.5in]{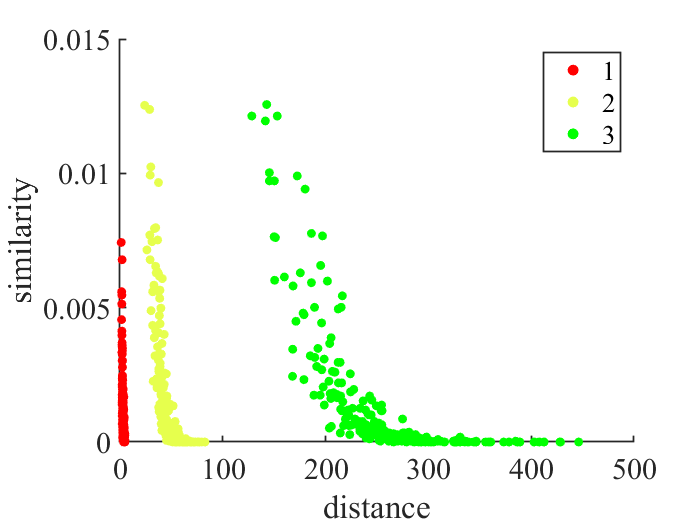}
   \caption{Isolation kernel with $\psi=250$}  
   \label{2b}
  \end{subfigure}  
\caption{Isolation kernel versus Gaussian kernel: Distributions of similarities of points  wrt the origin for three clusters of $N[0,1]$, $N[0,16]$ and $N[0,81]$ in different subspaces shown in Table \ref{v0}, where each is a 10-dimensional cluster (See the details in Footnote 1.) The similarities are  computed in the 50-dimensional space. The left-most point in each cluster is the point closest to the origin $O$ having the highest similarity: $x_{k1}$ is the red left-most point; $x_{k2}$ is the yellow (or green) left-most point.}  
    \label{c1}  
\end{figure}

An example distribution of similarities based on the dataset shown in Table \ref{v0} is given in Figure~\ref{c1}. Let $x_{k1}$ be the origin $O$'s closest point in the dense cluster (i.e., cluster 1); and $x_{k2}$ be $O$'s closest point in a sparse cluster (cluster 2 or 3). Figure \ref{2b}  shows that $K_\psi(O, x_{k1}) \approx K_\psi(O, x_{k2})$ when the Isolation kernel is used. When the Gaussian kernel is used,  $\mathcal{K}(O, x_{k1}) \gg \mathcal{K}(O, x_{k2})$, as shown in Figure \ref{2a}.

This explains why the points in the intersection are better mapped in the low-dimensional space by using the Isolation kernel than using the Gaussian kernel. %The same applies to the case shown in Table~\ref{v01}. \textcolor{red}{The last sentence is incorrect.}

In other words, the Isolation kernel ensures that the local structure is truly reflected in the similarities among local points  in the high-dimensional space, unlike the misrepresentation exhibited in Table \ref{v0} and Table \ref{v03} when the Gaussian kernel is used. As a result, t-SNE using the Isolation kernel produces the improved visualisation quality which has no misrepresentations.
 
\subsection{The second deficiency}
\subsubsection{Low computational efficiency problem with Gaussian kernel}

The use of a Gaussian kernel necessitates the search for a local bandwidth for each local point. t-SNE utilises a binary search for the value of $\sigma_i$ that makes the entropy of the distribution over neighbours equal to $\log K$, where $K$ is the effective number of local neighbours or ``perplexity'' \cite{TSNE}. %Another view is: adjust all bandwidths such that all $i$ have the same entropy: $\log(K) = -\sum_{j}p_{j|i} \ log_{2} \ p_{j|i} $.  
This search is the key component that determines the success or failure of t-SNE. A gradient descent search has been used successfully to perform the search for $n$ parameters for small datasets \cite{TSNE}. This formulation has two key limitations for large datasets. First, the need for $n$-parameters search poses a real limitation in terms of finding appropriate settings for a large number of parameters. Second, it cannot deal with large datasets because its low computational efficiency, i.e., the time complexity is $O(n^2)$.

\subsubsection{High computational efficiency with Isolation Kernel}
The computational complexities of the Guassian kernel and Isolation kernel \cite{ting2018IsolationKernel,aNNE} used in t-SNE are shown in Table \ref{complexity}.\footnote{Isolation kernel is derived from $t$ Voronoi diagrams and each Voronoi diagram has $\psi$ cells. Note that each Voronoi diagram does not need to be generated explicitly. This is because the cell into which a point falls can be determined by simply finding its nearest neighbor from the $\psi$ points.  This costs $O(t\psi)$. 
%Then calculating the similarity between any two points will determine the number of cells both points belong to, which costs  $O(2t\psi)$. 
Therefore, the time complexity of calculating pairwise similarity for $m$ points cost $O(t\psi m^2)$.} Although the parameter $\psi$ of Isolation kernel corresponds to the bandwidth parameter of the Gaussian kernel, the Isolation kernel needs no optimisation to determine $n$ bandwidths locally. This is because the partitioning mechanism used by the Isolation kernel produces small partitions in  dense regions and large partitions in  sparse regions; and the sizes of the partitions are monotonically decreasing with respect to $\psi$. As the local adaptation has already been done during the process of deriving the kernel, no further adaptation is required after the kernel is derived. 

Though the Isolation kernel derivation from data takes constant $O(t\psi)$ time, it is significantly less than the optimisation required to determine $n$ bandwidths which takes $O(n^2)$ time in Gaussian kernel. For a large dataset, when using Gaussian kernel, it is infeasible to estimate a large number of bandwidths with an appropriate degree of accuracy, and its computational cost is prohibitively high. In contrast, the consequence of using Isolation kernel is that the  runtime of step 1 in the t-SNE algorithm is significantly reduced. Thus, the Isolation kernel enables t-SNE to deal with large datasets. More experimental details are provided in Sections \ref{sec_large_datasets} and \ref{speedup}.

\begin{table}[t]
\setlength{\tabcolsep}{3pt}
\centering
\caption{Time complexities of t-SNE in steps (1) kernel building, (2) computing the similarity, and (3) mapping from high to low dimensions. $r$ is the number of iterations used for bandwidth search for the Gaussian kernel; and $s$ is the number of iterations in t-SNE mapping. $m (\le n)$ is the subsample size used for the mapping. For small datasets: $m=n$.}

\begin{tabular}{|l|c|c|}
\hline  
 & Gaussian kernel & Isolation kernel \\
\hline 
Step 1: Kernel building & $\mathcal{O}(rn^{2})$ & $\mathcal{O}(t\psi)$ \\ 
Step 2: Matrix calculation  & $\mathcal{O}(m^{2})$ & $\mathcal{O}(t\psi  m^2)$  \\ 
\hdashline
Step 3: t-SNE Mapping & \multicolumn{2}{c|}{$\mathcal{O}(sm^{2})$} \\ 
\hline 
\end{tabular} 
\label{complexity}
\end{table}

\section{The proposed solution: using the Isolation kernel in t-SNE}
\label{sec_IK1}

Since t-SNE needs a data-dependent kernel,
we propose to use a recent data-dependent kernel called Isolation kernel \cite{ting2018IsolationKernel,aNNE} to replace the data-independent Gaussian kernel in t-SNE. 

The Isolation kernel is a perfect match for the task because a data-dependent kernel, by definition, adapts to local 
distribution without any additional optimisation. The kernel replacement is conducted in the component in the high-dimensional space only, leaving the other components of the t-SNE procedure unchanged. 

Sections \ref{sec_IK} and \ref{sec_IK-vs-GK} are literature reviews of Isolation Kernel \cite{ting2018IsolationKernel} and a known fact \cite{aNNE}. Sections \ref{sec_small_samples}, \ref{sec_well-defined} and \ref{sec_IK-t-SNE} are our original contributions to Isolation kernel and t-SNE in this paper. 

\subsection{Isolation kernel}
\label{sec_IK}
The key idea of Isolation kernel is that using a space partitioning strategy to split the data space into different cells, e.g., we uniformly sample $\psi$ points from the given dataset and generate $\psi$ Voronoi cells, then the similarity between any two points is how likely the two points can be split into the same cell.  

The  details of Isolation kernel \cite{ting2018IsolationKernel,aNNE} are provided below.
 
 Let $D=\{{ x}_1,\dots,{ x}_n\}, { x}_i \in \mathbb{R}^d$ be a dataset sampled from an unknown probability density function ${ x}_i \sim F$. Moreover, 
let $\mathds{H}_\psi(D)$ denote the set of
all partitionings $H$ admissible for the given dataset $D$, 
where each $H$ covers the entire space of $\mathbb{R}^d$; and each of the $\psi$ isolating partitions $\theta[{ z}] \in H$ isolates one data point ${ z}$ from the rest of the points in a random subset $\mathcal D \subset D$, and $|\mathcal D|=\psi$. In our implementation, $H$ is a Voronoi diagram generated from $\mathcal D$.

\begin{definition} For any two points ${ x}, { y} \in \mathbb{R}^d$,
	the Isolation kernel of ${ x}$ and ${ y}$ wrt $D$ is defined to be the expectation taken over the probability distribution on all partitionings $H \in \mathds{H}_\psi(D)$ that both ${ x}$ and ${ y}$  fall into the same isolating partition $\theta[{ z}] \in H, { z} \in \mathcal{D}$:
	\begin{eqnarray} 
K_\psi({ x},{ y}|D) &=&  {\mathbb E}_{\mathds{H}_\psi(D)} [\mathds{1}({ x},{ y} \in \theta[{ z}]\ | \ \theta[{ z}] \in H)] %\nonumber \\
%&=& {\mathbb E}_{\mathcal{D} \subset D} [\mathds{1}({ x},{ y}\in \theta[{ z}]\ | \ { z}\in \mathcal{D})]  \nonumber\\
%&=& P({ x},{ y}\in \theta[{ z}]\ | \ { z}\in \mathcal{D} \subset D)
		\label{eqn_kernel}
	\end{eqnarray}
where $\mathds{1}(\cdot)$ is an indicator function.
\end{definition}

In practice, the Isolation kernel $K_\psi$ is constructed using a finite number of partitionings $H_i, i=1,\dots,t$, where each $H_i$ is created using $\mathcal{D}_i \subset D$:
\begin{eqnarray}
K_\psi({ x},{ y}|D)  & = &  \frac{1}{t} \sum_{i=1}^t   \mathds{1}({ x},{ y} \in \theta\ | \ \theta \in H_i) \nonumber\\
 & = & \frac{1}{t} \sum_{i=1}^t \sum_{\theta \in H_i}   \mathds{1}({ x}\in \theta)\mathds{1}({ y}\in \theta) 
 \label{Eqn_IK}
\end{eqnarray}

\noindent where $\theta$ is a shorthand for $\theta[{ z}]$; and $t$ can usually be set to a default value. $\psi$ is the sharpness parameter and the only parameter of the Isolation kernel. The larger $\psi$ is, the sharper the kernel distribution is. This corresponds to $\sigma$ in the Gaussian kernel, i.e., the smaller $\sigma$ is, the narrower the kernel distribution is. Note that $t$ is the number of partitionings and $t$ can be fixed to a large value to ensure the stability of the estimation.

As Equation (\ref{Eqn_IK}) is quadratic, $K_\psi$ is  a valid kernel. For brevity, $K_\psi({ x},{ y})$ is used to denote $K_\psi({ x},{ y}|D)$ hereafter.

\subsection{How Isolation kernel differs from Gaussian kernel}
\label{sec_IK-vs-GK}
The key difference is that the Isolation kernel adapts to local density distribution, but the Gaussian kernel is independent of the data distribution.

In addition, the technical differences can be observed in two aspects. First, the Isolation kernel has no closed-form expression. Second, it is derived directly from a dataset, without explicit learning or optimisation. Its adaptation to local density is a direct outcome of its isolation mechanism used to partition space, i.e., the mechanism produces large partitions in sparse regions and small partitions in dense regions \cite{ting2018IsolationKernel,aNNE}. A natural isolation mechanism that has this characteristic is a Voronoi diagram. Given a sample of the underlining distribution, each Voronoi cell isolates a point from the rest of the points in the sample; and the cells are small in the dense region and large in the sparse region. Note that the Voronoi diagram is obtained very efficiently, i.e.,   given a sample, nothing else needs to be done in the training stage because boundaries in the Voronoi diagram can be obtained at the testing stage as the equal distance between the two nearest points in the given sample. 

\begin{figure}[!bt]
\centering
  \begin{subfigure}{.49\textwidth}
  \centering
    \includegraphics[width=1.6in]{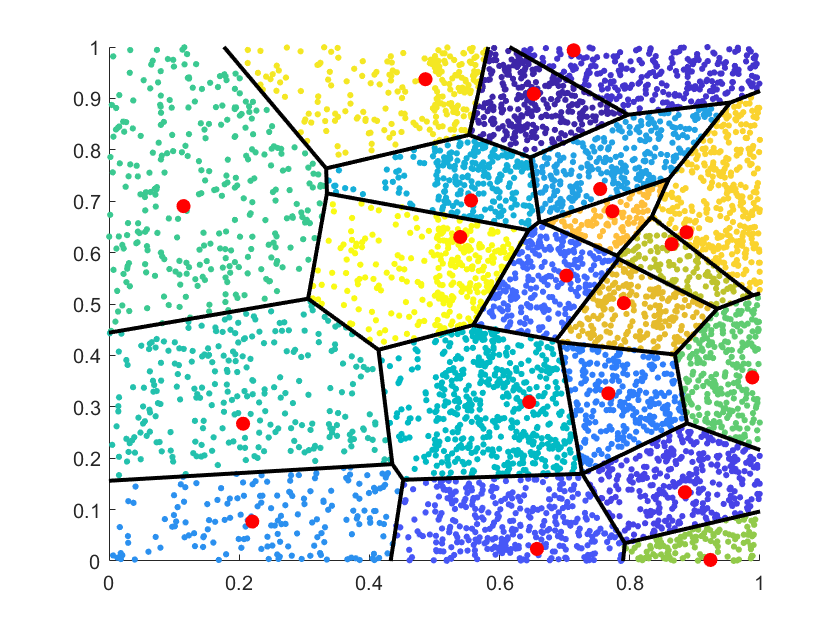}
  \caption{$\psi=16$}
  \end{subfigure}  %
  \begin{subfigure}{.49\textwidth}
  \centering
    \includegraphics[width=1.6in]{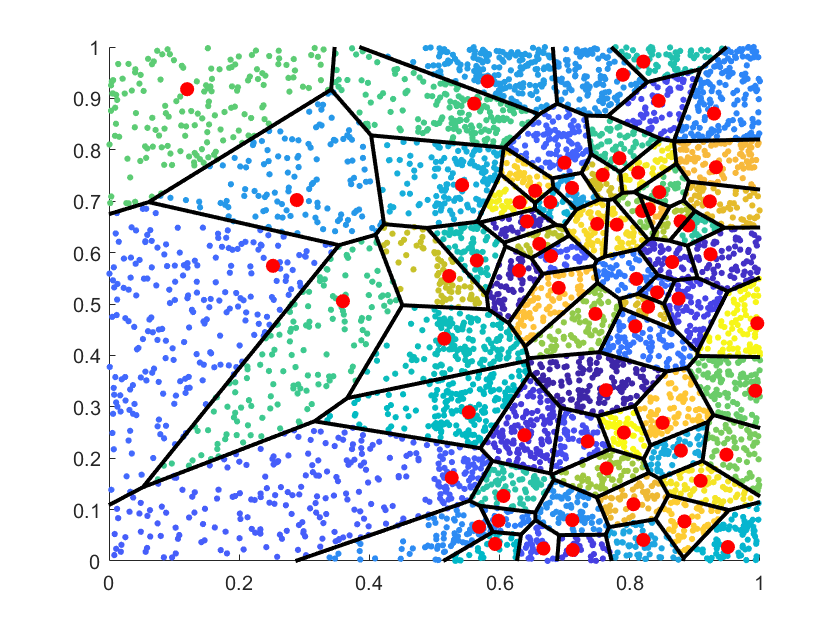}
  \caption{$\psi=64$}
  \end{subfigure}
\caption{Two examples of partitioning $H$ using the nearest neighbour (a Voronoi diagram) on a dataset having two regions of uniform densities, where the left half has a lower density than the right half.}
    \label{ill}  
\end{figure}

Figure \ref{ill} shows two examples of partitioning $H$ using the nearest neighbour or a Voronoi diagram on the same dataset with two different subsample sizes $\psi$. These examples show that there are more (small) cells in the dense region than (large) cells in the sparse region for each $\psi$; and the sizes of the cells are usually decreasing with respect to $\psi$. Two points located in the same cell get the similarity score of 1 in a partitioning. The final Isolation kernel similarity between two points is the probability of both points falling into the same cell over a finite number of partitionings, as shown in Equation \ref{Eqn_IK}. Examples of kernel distribution due to different $\psi$ values are shown in Appendix B, so as the implementation details.

\subsection{The Isolation kernel makes full use of the distributional information in small samples}
\label{sec_small_samples}

%The second technical difference described in the last section reveals the value of small samples and how it could be extracted without a computationally expensive process.

The Isolation kernel only requires  small samples ($\psi$) for the space partitioning without a computationally expensive process.

A small sample of a dataset contains data distributional information which is sufficient to build a data-dependent kernel. The Isolation kernel extracts this information in the form of a Voronoi diagram, which depicts the relative densities between regions.

In contrast, using a data-independent measure such as the Gaussian kernel, the distributional information in a dataset is ignored and each point in the input space is treated as an independent point. In order to get the distributional information in the form of variable bandwidths that are adaptive to the local distribution, a separate optimisation process is required, as conducted in step 1 of the t-SNE algorithm.

It is important to note that when they could not handle a large dataset, most methods may use small samples as a mitigation approach, and this inevitably trades off runtime with accuracy. But it is not the case for the Isolation kernel where small samples are the key in achieving high accuracy; and samples larger than the optimal $\psi$ will degrade the accuracy of Isolation kernel. See further discussion on this issue in Section \ref{sec_discussion}.

In other words, by using the Gaussian kernel, t-SNE must employ a computationally expensive approach to get the distributional information in a dataset. It does not exploit the same information which is freely available in small samples of the dataset. The Isolation kernel is a direct approach that makes full use of the distributional information freely available in small samples of a dataset.

\subsection{The Isolation kernel is well-defined}
\label{sec_well-defined}
The Isolation kernel has the following well-defined data-dependent characteristic: {\bf two points in a sparse region are more similar than two points of equal inter-point distance in a dense region} \cite{ting2018IsolationKernel}. 

Using a specific implementation of Isolation kernel (see Appendix B), \cite{aNNE} have provided the following Lemma (see its proof in their paper):

\begin{lemma}
\cite{aNNE} $\forall x_i, x_j \in \mathcal{X}_\mathsf{S}$ (sparse region) and $\forall x_k,x_\ell \in \mathcal{X}_\mathsf{T}$ (dense region) such that $\forall_{y\in \mathcal{X}_\mathsf{S}, z\in \mathcal{X}_\mathsf{T}} \ \rho(y)<\rho(z)$,
the nearest neighbour-induced Isolation kernel $K_\psi$ has the characteristic that for $\parallel x_i - x_j \parallel\ =\ \parallel x_k - x_\ell \parallel$ implies
\begin{eqnarray}
% P(x,y\in \theta[z]) > P(x',y'\in \theta[z'])  \equiv \hspace{3cm} \nonumber \\ 
K_\psi( x_i, x_j) >    K_\psi( x_k,x_\ell)
\label{eqn_condition}
\end{eqnarray}
where 
$\parallel x-y\parallel$ is the distance between $x$ and $y$; and $\rho(x)$ denotes the density at point $x$.
\label{lemma_dependent}
\end{lemma}

Let $p_{b|a}$ be the probability that $x_a$ would pick $x_b$ as its neighbour. 

We provide two corollaries from Lemma 1 as follows.

\begin{corollary} $x_i$ is more likely to pick $x_j$ as a neighbour than $x_k$ is to pick $x_\ell$ as a neighbour, i.e., $p_{j|i} > p_{\ell|k}$ for $\forall_{a,b}\ p_{b|a} \propto K_\psi(x_a, x_b)$.
 
\label{cor_1}
\end{corollary}
 
This is because $x_k$ in the dense region is more likely to pick a point closer than $x_\ell$ as its neighbour, in comparison with $x_i$ picking $x_j$ as a neighbour in the sparse region, given that $\parallel x_i - x_j \parallel\ =\ \parallel x_k - x_\ell \parallel$.

\begin{corollary} $\forall_{a,b}\ p_{b|a} \propto \frac{1}{\bar{\rho}(\mathcal{X}_\mathsf{A})}$,  where $x_a, x_b \in \mathcal{X}_\mathsf{A}$ is a region in $\mathcal{X}$; and $\bar{\rho}$ is an average density of a region.
\label{cor_2}
\end{corollary}

Using a data-dependent kernel with a well-defined characteristic as specified in Lemma~1, we can establish that the probability that $x_a$ would pick $x_b$, $p_{b|a}$, is inversely proportional to the density of the local region.

This becomes the basis in setting a reference probability in the high-dimensional space.

It is interesting to note that the adaptation of Gaussian kernel by optimising $n$ bandwidths attempts to achieve a similar outcome, as stipulated in Corollaries \ref{cor_1} and \ref{cor_2}. Yet, it is unclear that a similar data-dependent characteristic, as stated in Lemma \ref{lemma_dependent}, can be formally stated for the adaptive Gaussian kernel. This is because the similarity cannot be computed for all $x \in \mathbb{R}^d$ (except those in the given dataset.)

\subsection{t-SNE with the Isolation kernel}
\label{sec_IK-t-SNE}
We propose to replace $\mathcal{K}$ with $K_\psi$ in defining $p_{j|i}$ in Equation (\ref{eqn_p}), i.e.,

\begin{equation}
p_{j|i} = \frac{K_\psi(x_i,x_j)}{\sum_{k \ne i} K_\psi(x_i,x_k)}.  
\label{eqn_p_IK}
\end{equation}

The rest of the procedure of t-SNE remains unchanged.

The procedure of t-SNE with the Isolation kernel is provided in Algorithm \ref{alg_IK-t-SNE}. 

Note that the only difference between the two algorithms is step 1; and Eq \ref{eqn_p_IK} (instead of Eq \ref{eqn_p}) in step 2.  

    \begin{algorithm}[!htbp]
        \caption{t-SNE$(D, \psi, m)$ which employs the Isolation kernel}
        \begin{algorithmic}[1]  
            \Require $D$ - Dataset $\{x_1,\dots,x_n \}$; $\psi$ - sharpness parameter of the Isolation kernel
        \State Build a space partitioning model using $t$ sets of $\psi$ data points  for the Isolation kernel $K_\psi$  
        \State Compute matrix $[p_{ij}]_{m \times m}$ according to Equations \ref{eqn_p2} \& \ref{eqn_p_IK}
%        \State Compute $p_{j|i}$ based on Isolation kernel $K_\psi(x_i,x_j)$
%        \State Set $p_{ij} = \frac{p_{j|i} + p_{i|j}}{2n}$
        \State Compute low-dimensional $D'$ and $p'_{ij}$ which minimise the KL divergence 
        \State Output low-dimensional data representation $D'=\{x'_1, \dots, x'_m\}$
        \end{algorithmic}
        \label{alg_IK-t-SNE}
    \end{algorithm}

\section{Empirical Evaluation}
\label{sec_evaluation}

This section presents the three evaluation methods we adopt, evaluation results, runtime comparison and a scalability test.

\subsection{Evaluation measures} %\label{app1}

We used a qualitative assessment $R(k)$ to evaluate the preservation of $k$-ary neighbourhoods \cite{lee2009qualityQNX,lee2013typeRNX,lee2015multi}, defined as follows:

\begin{equation}
R(k)=\frac{(n-1)Q(k)-k}{n-1-k}
\label{RNX}
\end{equation}

\noindent where $Q(k)=\sum_{i=1}^{n}\frac{1}{nk}|kNN(x_i) \cap kNN(x'_i)|$

\noindent
and $kNN(x)$ is the set of $k$ nearest neighbours of $x$; and $x'$ is the corresponding low-dimensional (LD) point of the high-dimensional (HD) point $x$.

$R(k)$ measures the $k$-ary neighbourhood agreement between the HD and corresponding LD spaces. $R(k) \in [0,1]$; and the higher the score is, the better the neighbourhoods preserved in the LD space. In our experiments, we recorded the assessment with $k\in \lbrace0.01n,0.03n,..., 0.99n\rbrace$ and produced the curve, i.e., $k$ vs $R(k)$.

To aggregate the performance over the different $k$-ary neighbourhoods, we calculate the area under the $R(k)$ curve in the log plot \cite{lee2013typeRNX} as:

\begin{equation}
AUC_{RNX}=\frac{\sum_{k}R(k)/k}{\sum_{k} 1/k}
\label{AUCRNX}
\end{equation}

AUC$_{RNX}$ assesses the average quality weighted by $k$, i.e., errors in large neighbourhoods with large $k$ contribute less than that with small $k$ to the average quality. 
 
In addition, the purpose of many methods of dimensionality reduction  is to identify HD clusters in the LD space such as in a 2-dimensional scatter plot. Since all the datasets we used for evaluation have ground truth (labels), we can use measures for clustering validation to evaluate whether all clusters can be correctly identified after they are projected into the LD space. Here we select two popular indices of cluster validation, i.e., Davies-Bouldin (DB) index \cite{davies1979clusterDB} and Calinski-Harabasz (CH) index \cite{calinski1974dendriteCH}. Their details are given as follows.

Let $x$ be an instance in a cluster $C_i$ which has $n_i$ instances with the centre as $c_i$. The Davies-Bouldin (DB) index can be obtained as% $S_i$ measures the scatter within the cluster $C_i$ as:

\begin{equation}
DB=\frac{1}{N_C}\sum_{i}max_{j,j\neq i}\{[\frac{1}{n_i}\sum_{x\in C_i}||x-c_i||_2+\frac{1}{n_j}\sum_{x\in C_j}||x-c_j||_2]/||c_i-c_j||_2\}
\label{DB}
\end{equation}

\noindent where $N_C$ is the number of clusters in the dataset.

Calinski-Harabasz (CH) index is calculated as

\begin{equation}
CH=\frac{\sum_i n_i||c_i-c||_2/(N_C-1)}{\sum_i\sum_{x\in C_i}||x-c_i||_2/(n-N_C)}
\label{CH}
\end{equation}

\noindent where $c$ is the centre of the dataset.

Both measures take the similarity of points within a cluster and the similarity between clusters into consideration, but in different ways. These measures assign the best score to the algorithm that produces clusters with low intra-cluster distances and high inter-cluster distances. Note that the higher the CH score, the better the cluster distribution; while the lower the DB score is, the better the cluster distribution is. 

All algorithms used in the following experiments were implemented in Matlab 2019b and were run on a machine with 14 cores (Intel Xeon E5-2690 v4 @ 2.59 GHz) and 256GB memory.\footnote{The Matlab code of 
t-SNE and the Isolation kernel are from \url{https://lvdmaaten.github.io/tsne} and \url{https://github.com/cswords/anne-dbscan-demo}, respectively. A demonstration of using t-SNE with Isolation kernel can be obtained from \url{https://github.com/zhuye88/IKt-sne}.} All datasets were normalised using the $min$-$max$ normalisation to yield each attribute to be in [0,1] before the experiments began. We also use the $min$-$max$ normalisation on the t-SNE results before calculating DB and CH scores. %The same normalisation was used on the projected datasets before calculating CH and DB scores.% \textcolor{red}{The last two sentences mean the same thing, right?}

\subsection{Evaluation results}

This section presents the result of utility evaluation of isolation kernel and Gaussian kernel in t-SNE using 21 real-world datasets\footnote{COIL20, HumanActivity and Isolet are from \cite {li2016feature}; News20 and Rcv1 are from  \cite{CC01a}; and all other real-world datasets are from UCI Machine Learning Repository \cite{Lichman:2013}.} with different data sizes and dimensions. We report the best performance of each algorithm with a systematic parameter search with the range shown in Table \ref{para}.\footnote{The search range used for t-SNE is significantly larger than that suggested in the t-SNE paper ``the performance of SNE is fairly robust to changes in the perplexity, and typical values are between 5 and 50'' \cite{TSNE}.} Note that there is only one manual parameter $\psi$ to control the partitioning mechanism, and the other parameter $t$ can be fixed to a default number.

\begin{table}[!bt]
  \centering
%   \footnotesize 
  \caption{Parameters and their search ranges for each kernel function.}
    \begin{tabular}{|c|c|}
    \hline
     & Parameters with search range \\
    \hline
    Gaussian kernel & $perplexity \in\lbrace 1, 5,...,97, 0.01n, 0.05n,..., 0.97n \rbrace$; $tolerance =0.00005$ \\         
    Isolation kernel& $\psi \in\lbrace  1, 5, ...,97, 0.01n,0.05n,..., 0.97n\rbrace$; $t=200$ \\  
    \hline
    \end{tabular}%
  \label{para}%
\end{table}%

  \begin{table}[t]
    \centering
  \renewcommand{\arraystretch}{1.}
   \setlength{\tabcolsep}{4.1pt}
    \caption{Evaluation results on real-world datasets. For each dataset, the best performer, GK (Gaussian kernel) or IK (Isolation kernel)  w.r.t. each evaluation measure is boldfaced. Note that the higher the $AUC_{RNX}$ and CH scores indicate the better a cluster distribution; while a lower DB score indicates a better cluster distribution.} 
    \label{tbl_results}
      \begin{tabular}{|l|r|r|rr|rr|rr|}
      \hline
      \multirow{3}[6]{*}{Dataset} & \multirow{3}[6]{*}{\#Points} & \multirow{3}[6]{*}{\#Attr} & \multicolumn{6}{c|}{Evaluation measure} \\
\cline{4-9}            &       &       & \multicolumn{2}{c|}{$AUC_{RNX}$} & \multicolumn{2}{c|}{DB} & \multicolumn{2}{c|}{CH} \\
\cline{4-9}            &       &       & GK    & IK    & GK    & IK    & GK    & IK \\
      \hline
    Wine  & 178   & 13    & 0.65  & \textbf{0.67} & 0.52  & \textbf{0.43} & 625   & \textbf{853} \\
    Dermatology & 358   & 34    & 0.68  & \textbf{0.684} & 0.47  & \textbf{0.38} & 3679  & \textbf{8171} \\
    ForestType & 523   & 27    & 0.70  & \textbf{0.71} & 0.91  & \textbf{0.87} & 467   & \textbf{620} \\
    WDBC  & 569   & 30    & 0.65  & \textbf{0.67} & 0.68  & \textbf{0.58} & 876   & \textbf{1167} \\
    ILPD  & 579   & 9     & 0.68  & \textbf{0.69} & 3.92  & \textbf{3.71} & 24    & \textbf{28} \\
    Control & 600   & 60    & 0.69  & \textbf{0.70} & 0.67  & \textbf{0.64} & 4011  & \textbf{6816} \\
    Pima  & 768   & 8     & 0.70  & \textbf{0.72} & 2.97  & \textbf{2.87} & 67    & \textbf{76} \\
    Parkinson & 1040  & 26    & 0.70  & \textbf{0.74} & 6.69  & \textbf{6.35} & 20    & \textbf{22} \\
    Biodeg & 1055  & 41    & 0.74  & \textbf{0.77} & 2.04  & \textbf{1.69} & 154   & \textbf{183} \\
    Mice  & 1080  & 83    & 0.79  & \textbf{0.82} & 0.32  & \textbf{0.15} & 8326  & \textbf{50067} \\
    Messidor & 1151  & 19    & 0.71  & \textbf{0.74} & 6.36  & \textbf{6.02} & 26    & \textbf{28} \\
    Hill  & 1212  & 100   & 0.69  & \textbf{0.73} & 16.71  & \textbf{15.10} & 4     & \textbf{4.5} \\
    COIL20 & 1440  & 1024  & 0.75  & \textbf{0.79} & 3.10  & \textbf{2.67} & 2352  & \textbf{3730} \\
    HumanActivity & 1492  & 561   & 0.78  & \textbf{0.79} & 2.68  & \textbf{2.29} & 1225  & \textbf{1728} \\
    Isolet & 1560  & 617   & 0.80  & \textbf{0.81} & 1.83  & \textbf{1.41} & 1746  & \textbf{2812} \\
    Segment & 2310  & 19    & 0.68  & \textbf{0.72} & \textbf{1.30} & 1.51  & 7363  & \textbf{8337} \\
    Spam  & 4601  & 57    & 0.67  & \textbf{0.70} & 1.36  & \textbf{1.33} & 1832  & \textbf{1874} \\
    News20 & 9998  & 1355191 & \textbf{0.27} & 0.25  & \textbf{1.57} & 1.92  & \textbf{2732} & 2320  \\
    MNIST & 10000 & 784   & \textbf{0.66} & 0.64  & 1.13  & \textbf{0.93} & 6452  & \textbf{8024} \\
    Rcv1  & 10121 & 47236 & \textbf{0.68} & 0.66  & 2.50  & \textbf{1.43} & 2421  & \textbf{4221} \\
    Pendig & 10992 & 16    & 0.69  & \textbf{0.693} & 1.14  & \textbf{1.10} & \textbf{6944} & 6777  \\
      \hline
      \multicolumn{3}{|r|}{\textit{Average}}   &  0.68  & 0.70  & 2.80  & 2.54  & 2445  & 5136     \\
      \hline
      \end{tabular}%
    \label{result}%
  \end{table}%
  
Table \ref{result} shows the results of the two kernels used in t-SNE. The Isolation kernel performs better on 18 out of 21 datasets in terms of $AUC_{RNX}$, which means that the Isolation kernel enables t-SNE to preserve the local neighbourhoods much better than the Gaussian kernel. With regard to the cluster quality, the Isolation kernel performs better than the Gaussian kernel on 18 out of 21 datasets in terms of both DB and CH. Notice that when the Gaussian kernel is better, the performance gaps are usually small in any of the three measures. Overall, the Isolation kernel is better than the Gaussian kernel on 16 out of 21 datasets in all three measures. The reverse is true on one dataset only, i.e., News20. The visualization result on New20binary indicates there are significant overlaps between the two clusters in this dataset. This is reflected in the $AUC_{RNX}$ results which are significantly less than a random assignment ($AUC_{RNX}=0.5$). The visualization result of News20 is shown in Appendix C.

On the COIL20 dataset, we have identified a structural misrepresentation issue with the Gaussian kernel, similar to the one shown in Table \ref{v03}.  
Table \ref{c1c} shows the five clusters where the Gaussian kernel has misrepresented  structures in the high-dimensional space. The 3-dimensional results denote that the Isolation kernel depicts a more nuanced structural relationship between the five clusters; whereas the Gaussian kernel depicts that they are disparate five clusters, shown in the second column in Table \ref{c1c}. Also, note that a reference point $\times$ is close to all five clusters when the Isolation kernel is used, but it is far from many clusters when a Gaussian kernel is used.

\begin{table} [!tb] 
\centering
  \begin{tabular}{cccc}
    %   \multicolumn{3}{c}{ \includegraphics[width=6in]{Fig/heat.png}} \\
    %   \multicolumn{3}{c}{Heatmap on the standard deviation of each attribute for five clusters}\\
    %   \\
    \hline
  & t-SNE in 2d  &     t-SNE on 5 selected classes in 3d  \\
      \hline 
    \quad   \begin{turn}{90}  \quad  Gaussian kernel    \end{turn} \quad \quad &       
       \includegraphics[width=2.5in]{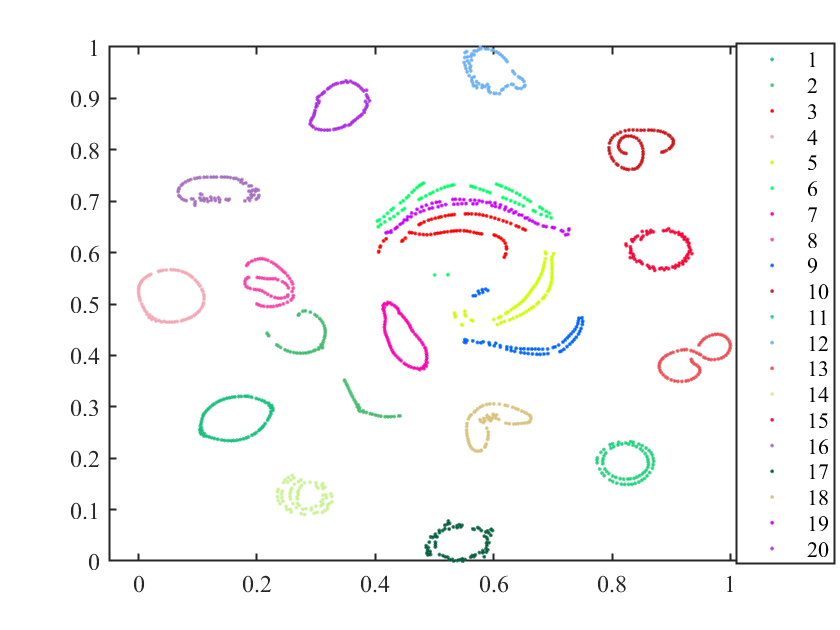}   &       
       \includegraphics[width=2.5in]{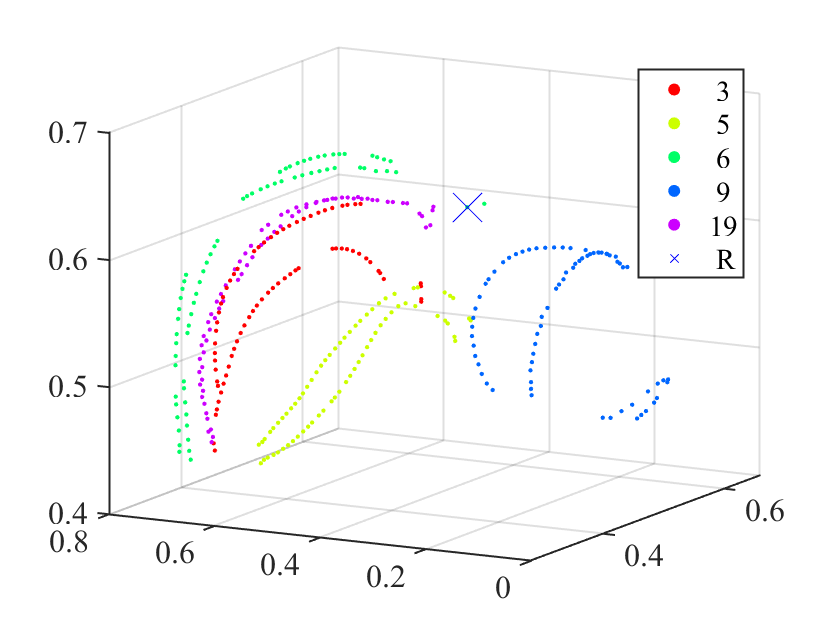}               \\  
       & (a) & (b) \\
       \begin{turn}{90} \quad Isolation kernel  \end{turn} &       
       \includegraphics[width=2.5in]{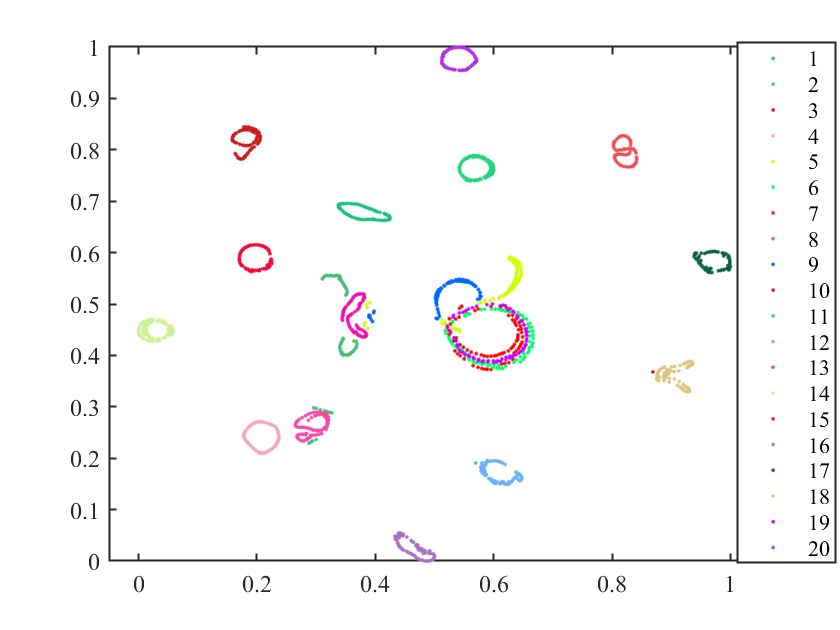}   &       
          \includegraphics[width=2.5in]{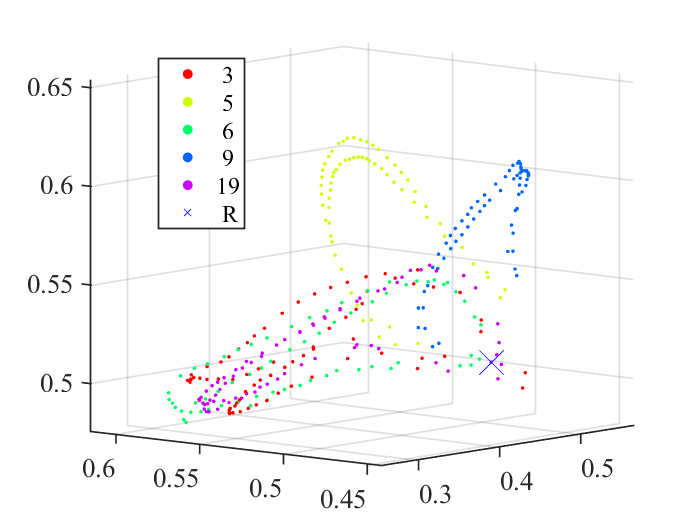}          \\  
               & (c) & (d) \\\hline
  \end{tabular}
  \caption{(a) and (c) show the t-SNE  visualisation results on COIL20 in a two-dimensional space. (b) and (d) show the five clusters and a reference point (indicated as $\times$ with the class label ``R'') on t-SNE visualisation results in a three-dimensional space.}%The Heatmap on the standard deviation of each attribute for the five cluster in shown on the top. }
 \label{c1c}  
\end{table}

\subsection{Runtime comparison}

Generally, both Gaussian Kernal and Isolation Kerner have quadratic time and space complexities. However, the Gaussian kernel in the original t-SNE needs a large number of iterations to search for the optimal local bandwidth for each point. as a result, the Gaussian kernel takes a much longer time in step 1 of the algorithm than the Isolation kernel.

\begin{figure}[ht]
\centering
  \begin{subfigure}{.49\textwidth}
  \centering
    \includegraphics[width=1.75in]{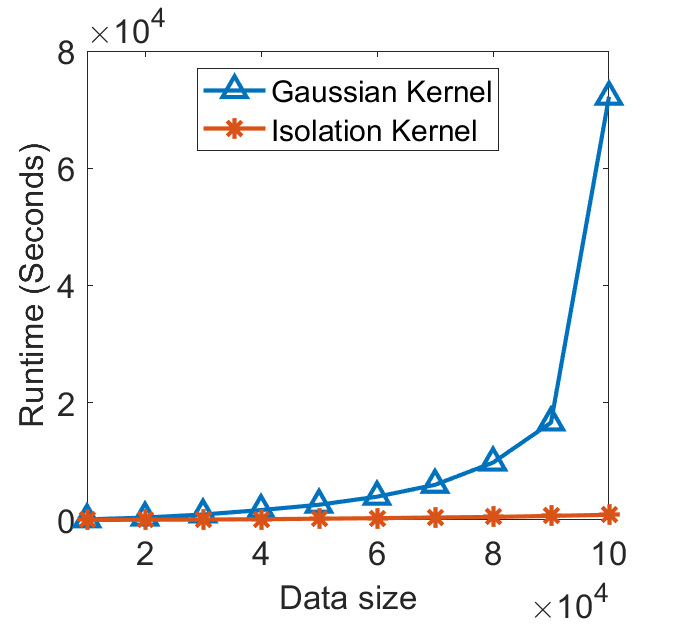}
    \caption{Runtime for Steps 1 \& 2 ($m=n$)}
  \end{subfigure}  %
  \begin{subfigure}{.49\textwidth}
  \centering
    \includegraphics[width=1.75in]{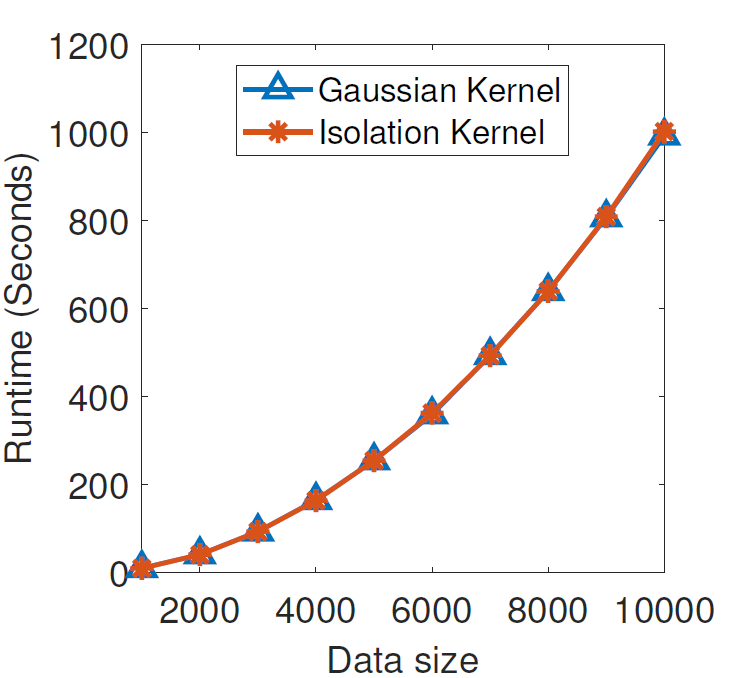}
  \caption{Runtime for Step 3}
  \end{subfigure}
\caption{CPU runtime comparison of Gaussian kernel and Isolation kernel used in t-SNE on a 2-dimensional synthetic dataset.}
    \label{runtime}  
\end{figure}

Figure \ref{runtime}  presents the two runtime comparisons of t-SNE with the two kernels on a synthetic dataset. Figure \ref{runtime}(a) shows that the Gaussian kernel is much slower than the Isolation kernel in similarity calculations. This is mainly due to the search required to tune the $n$ bandwidths in step 1 of the algorithm. It is interesting to note that though both similarities have $n^2$ time complexity, the constant is significantly lower in the Isolation kernel: if the data size is increased 10 times from 10,000 to 100,000, the Gaussian kernel increases its runtime 685 times; whereas the Isolation kernel increases 91 times only. As a result, with a dataset of 100,000 data points, the Isolation kernel\footnote{In addition, the Isolation kernel is amenable to GPU acceleration \cite{aNNE}. Our experiment shows that the runtime of Isolation kernel can be sped up by two orders of magnitude with a GPU machine, e.g., from 54 CPU seconds to 0.24 GPU seconds for a dataset of 25,000 data points.} is two orders of magnitude faster than the Gaussian kernel (887 seconds versus 72,196 seconds).

Figure \ref{runtime}(b) shows the runtime  of the mapping process in step 3 of Algorithms 1 and 2 which is the same for both algorithms. It is not surprising that their runtime are about the same in this step, regardless of the kernel employed.

Table \ref{runtime2} compared the CPU runtime of Gaussian kernel and Isolation kernel used in t-SNE on four real-world datasets. The t-SNE with the Isolation kernel is up to one order of magnitude faster than the t-SNE with Gaussian kernel in the first two steps. 
 
  \begin{table}[htbp]
    \centering
    \caption{CPU runtime (seconds) of t-SNE on four real-world datasets.}
      \begin{tabular}{|c|cc|cc|}
      \hline
            & \multicolumn{2}{c|}{Gaussian kernel} & \multicolumn{2}{c|}{Isolation kernel}  \\
\cline{2-5}            & Steps 1 \& 2 & Step 3 & Steps 1 \& 2 & Step 3  \\
      \hline
%      WDBC  & 0.45  & 4.76  & 0.12  & 4.64  \\
%      Mice  & 0.99  & 12.25 & 0.20   & 11.88 \\
%      Spam  & 20.18 & 203.84 & 2.25  & 202.1 \\
%      Pendig & 35.82 & 1146.64 & 12.17 & 1147.36 \\
      WDBC  & 0.5  & 4.8  & 0.1  & 4.6  \\
      Mice  & 1.0  & 12.3 & 0.2   & 11.9 \\
      Spam  & 20.1 & 203.8 & 2.3  & 202.1 \\
      Pendig & 35.8 & 1146.6 & 12.2 & 1147.4 \\
      \hline
      \end{tabular}%
    \label{runtime2}%
  \end{table}% 

\subsection{Scalability testing}
\label{sec_large_datasets}

Here we show that the Isolation kernel enables t-SNE to deal with large datasets because step 1  takes constant time (once the parameters are fixed), rather than  $n^2$  when a Gaussian kernel is used.

This allows t-SNE to deal with a dataset with millions of data points in step 1, while using a subsample in steps 2 \& 3 to visualise the dataset in a low-dimensional space.

To demonstrate this ability, we use the MNIST8M dataset \cite{loosli-canu-bottou-2006} with 8.1 million points in step 1; and then use either the MNIST dataset or a subsample of 10,000 data points from MNIST8M in steps 2 \& 3 of t-SNE. The results of t-SNE with the Isolation kernel are shown in the last two columns in Table \ref{scale}. The results show that IK can get good CH scores with small $\psi$ values. It took 334s ($\psi=2048$) in steps 1 and 2, and 972s in step 3. Note that t-SNE with Gaussian kernel cannot be directly applied on this massive dataset in the same manner because it would take too long to complete step 1, as shown in Figure \ref{runtime}(a). 

\begin{table} [!tb] 
\setlength{\tabcolsep}{.5pt}
\centering
  \begin{tabular}{ccccc}
    \hline
  &  MNIST in all steps  &   MNIST8M (step 1)  & MNIST8M(step 1)    \\
      &  & \& MNIST (steps 2, 3) &  \& 10,000 (steps 2, 3)\\
      \hline 
       \begin{turn}{90}  \ \   Gaussian kernel    \end{turn} &       
       \includegraphics[width=1.56in]{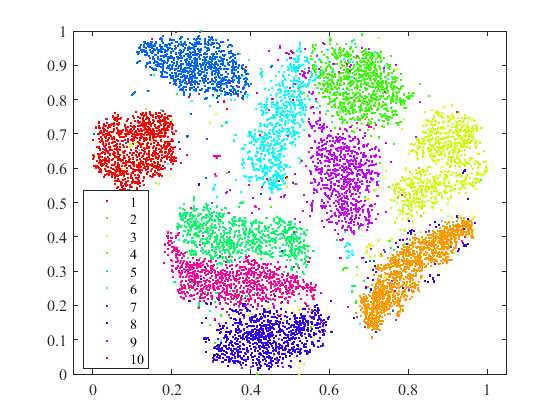}   &       
       \includegraphics[width=1.55in]{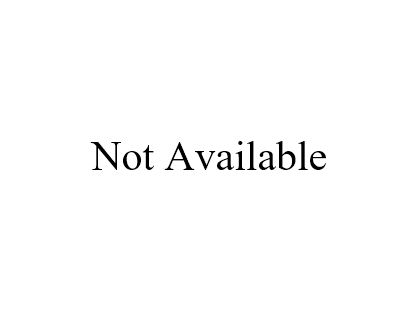}      &  \includegraphics[width=1.55in]{Fig/NA.png}      \\  
       & CH=6452 &   & \\
       \begin{turn}{90} \ \ Isolation kernel  \end{turn} &       
       \includegraphics[width=1.55in]{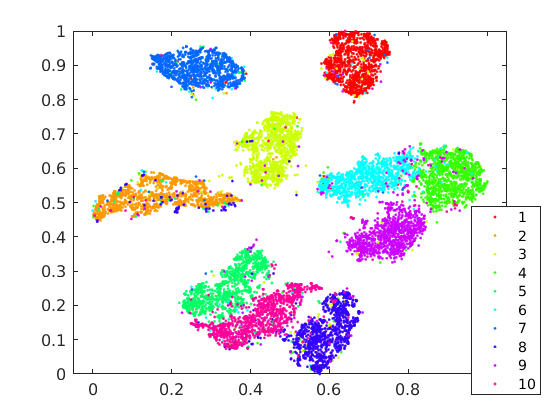}   &       
          \includegraphics[width=1.55in]{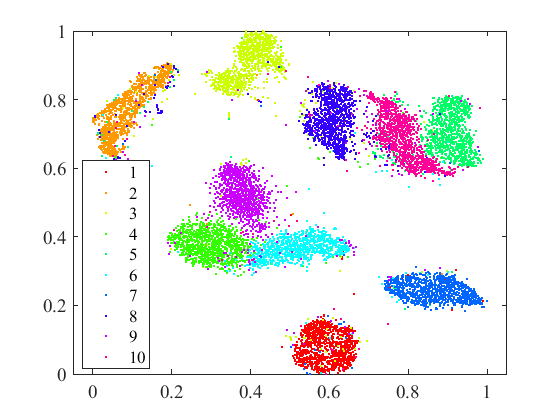}      &   \includegraphics[width=1.55in]{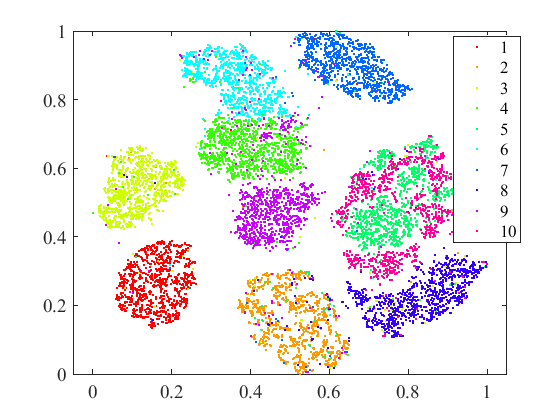}      \\  
             &CH=8024  & CH=8846 & CH=6335 \\\hline
  \end{tabular}
  \caption{t-SNE visualisation results on the MNIST and MNIST8M datasets.}
 \label{scale}  
\end{table}

The use of a subsample in steps 2 and 3 was previously suggested by \cite{TSNE}. However, the suggestion was to replace the Gaussian kernel with a graph similarity that employs a random walk method. This graph similarity approach has the same limitation as the Gaussian kernel because of its high time complexity. It requires a neighbourhood graph to be generated before a random walk kernel (or any graph kernel) can be used to measure similarities. While many graph kernels (see e.g., \cite{GraphKernel-Survey2020}) may be applied here, the key obstacle is the generation of the neighbourhood graph which has at least $O(n^2)$ time complexity.

In summary, employing Isolation kernel is the only method that takes constant time in step 1. Meanwhile, subsampling in step 2 and 3 enables t-SNE to process large-scale datasets without compromising the reference probability that needs to be established in step 1.

\section{Discussion}
\label{sec_discussion}

\subsection{The proposed method can benefit existing variants of t-SNE}
The common feature of existing variants of t-SNE is that they all use the Gaussian kernel.\footnote{There are other data-dependent kernels that can be used in t-SNE, see Appendix D for details.} The proposed idea can be applied to variants of stochastic neighbour embedding, e.g., NeRV \cite{venna2010information} and JSE \cite{lee2013typeRNX}, since they employ the same algorithm procedure as t-SNE. The only difference is the use of variants of cost function, i.e., type 1 or type 2 mixture of KL divergences.

In addition, Isolation kernel can be used in existing methods which aims to speed up t-SNE in step 3 of the algorithm. This is discussed in   Section \ref{speedup}.

\subsection{Isolation kernel performs optimally with small samples}

The finding---small samples (as the $\psi$ value) have better visualisation results than large samples---was formally analysed in the context of nearest neighbour anomaly detection \cite{LearningCurve-MLJ2017}. The work is motivated by the previous finding that small samples can produce better detection accuracy for some anomaly detectors than large samples (e.g., \cite{liu2008isolation,Sugiyama}.) The theoretical analysis based on computational geometry reveals that the geometry of data distribution has a direct impact on the sample size setting which is essential to  produce an optimal nearest neighbour anomaly detector \cite{LearningCurve-MLJ2017}. In a simple geometry such as a Gaussian distribution, a sample size of one data point (at the mean of Gaussian distribution) yields the optimal nearest neighbour anomaly detector; and a sample of more data points will produce a worse performing detector. In a more complex geometry of data distribution (e.g., a mixture of multiple Gaussian distributions), while the optimal sample size is more than one data point, a sample size over the optimal one also produces a worse performing detector. See \cite{LearningCurve-MLJ2017} for details.
 
The above result can  explain the effect of small samples in  Isolation kernel described in Section \ref{sec_small_samples}: the optimal sample size is the representative sample for the underlying geometry of data distribution, allowing the Isolation kernel to model relative similarities between different regions most effectively.

In summary, most methods use small samples as a compromising approach when failing to handle large datasets. It comes at the cost of low accuracy and longer runtime. However, algorithms employing Isolation kernel can process large datasets without trading off accuracy and efficiency due to the resultant sample. While $\psi$ of the Isolation kernel serves the primary purpose of a kernel parameter like the bandwidth parameter of Gaussian kernel, the resultant sample size enables algorithms that employ the Isolation kernel to deal with large datasets without compromising the accuracy of the task.     

\subsection{Methods to speed up t-SNE}
\label{speedup}
Scalability is an open issue for applying unsupervised distance metric learning approaches on large datasets \cite{wang2015survey}. As mentioned before, currently, there are two ways to speed up t-SNE: subsampling (which is a mitigation approach discussed in Section \ref{sec_small_samples}), and another is via some approximation to reduce runtime in step 3. 
 
The two approximation methods mentioned in the literature review are (i) the Barnes-Hut algorithm in conjunction with the dual-tree algorithm \cite{van2014accelerating}, and (ii) interpolating onto an equispaced  grid in order to use the fast Fourier transform to perform the convolution required in step 3 of the t-SNE algorithm \cite{linderman2019fast}. However, these approximation methods sacrifice accuracy for efficiency. For example, opt-SNE \cite{belkina2019automated}  utilises Kullback-Leibler divergence evaluation to automatically identify the tailored parameters in the optimisation procedure of t-SNE, in order to reduce the iteration time and improve the embedding quality. Nevertheless, all of these methods are still based on Gaussian kernel. Therefore, they still have the same deficiency of misrepresented structures as the original t-SNE, as discussed in Section \ref{sec_pint-based_Bandwidth}. Appendix E and Appendix F show examples of these outcomes of  FIt-SNE \cite{linderman2019fast} and opt-SNE \cite{belkina2019automated}, respectively.% 

In a nutshell, the proposed method of using Isolation kernel in t-SNE offers (i) the only way to establish the reference probability in step 1 using a large dataset (without parallelisation); and (ii)  a way to speed up t-SNE, which is an alternative to existing speedup methods. The use of a subsample, as a mitigation approach, in step 1 compromises the accuracy of reference probability. The use of an approximation method in step 3 reduces the quality of the dimensionality reduction. These existing methods in speeding up t-SNE still employ Gaussian kernel; and thus they fail to address the two deficiencies we have identified.
 
%\newpage
\section{Conclusions}

This paper identifies two deficiencies in t-SNE due to the use of Gaussian kernel. First, the point-based-bandwidth Gaussian kernel often creates misrepresented structure(s) which do not exist in the given dataset under some conditions. Second, the data-independent Gaussian kernel largely increases the computation load resulted from the need in determining $n$ bandwidths for a dataset of $n$ points and thus unable to deal with large datasets. Though some methods have been suggested to trade off accuracy for faster running speed, the underlying issue due to the use of Gaussian kernel remains unresolved.
 
Since the root cause of these deficiencies is the use of a data-independent kernel, we propose to simply replace Gaussian kernel with a data-dependent kernel called Isolation kernel.

We show that the use of Isolation kernel in t-SNE overcomes the drawback of misrepresenting some structures in the data, which often occurs when Gaussian kernel is applied in t-SNE. Also, the use of Isolation kernel yields a more efficient similarity computation because {\em data-dependent} Isolation kernel has only one parameter that needs to be tuned. Unlike the existing methods in speeding up t-SNE, this efficient feature of Isolation kernel enables t-SNE to deal with large-scale datasets without trading off accuracy.

\section*{Appendix A. Visualisation results of t-SNE on subspace clusters having some shared attributes}  

 Here we use a dataset with three subspace clusters where all clusters share two same attributes only.
 The three clusters have the same Gaussian distribution $N[0,1]$. Cluster 1 has 500 points with relevant attributes from \#1 to \#51 dimensions; cluster 2 has 500 points with relevant attributes from \#50 to \#100 dimensions; and cluster 3 has 20 points with relevant attributes from \#50 to \#51 dimensions. All irrelevant  attributes of each cluster having  zero  values. Because most attributes of cluster 3 are zero, the overall distance between cluster 3 and cluster 1 or cluster 2 is much smaller than the distance between cluster 1 and cluster 2.

 Table \ref{sub1} shows the visualisation results of t-SNE with Gaussian kernel and Isolation kernel on the above-mentioned 100-dimensional dataset. It can be seen from the table that the Isolation kernel with small $\psi$ values presents the cluster structure correctly, i.e., the third cluster is in the centre and close to clusters 1 and 2. 
 
 In contrast, t-SNE with Gaussian kernel using $perplexity=50$ shows a small gap between clusters 1 and 2; the separation between cluster 3 and clusters 1 \& 2 are not clear. If increasing $perplexity$ to 250, three points that are close to the origin from cluster 3 (including the origin) become far away from clusters 1 and 2. This is because they got much smaller bandwidths than all other points due to the high density around the origin. As a result, they are very dissimilar to most other points.% \textcolor{red}{What happen to the origin? Why the Origin is not far away from the rest of the points?}

 \begin{table} [!htbp] 
 %\vspace{10mm}
  % \renewcommand{\arraystretch}{1.2}
 \setlength{\tabcolsep}{0pt}
 \centering
 \caption{Visualisation results of t-SNE with Gaussian kernel and Isolation kernel on a 100-dimensional dataset with three subspace clusters. Note that in (c), three points (including the origin) from cluster 3 are far away from clusters 1 and 2, as indicated with the red arrows.} 
   \begin{tabular}{cccc}
     \hline 
       \begin{turn}{90} \ \ Gaussian kernel \end{turn}&      \includegraphics[width=1.6in]{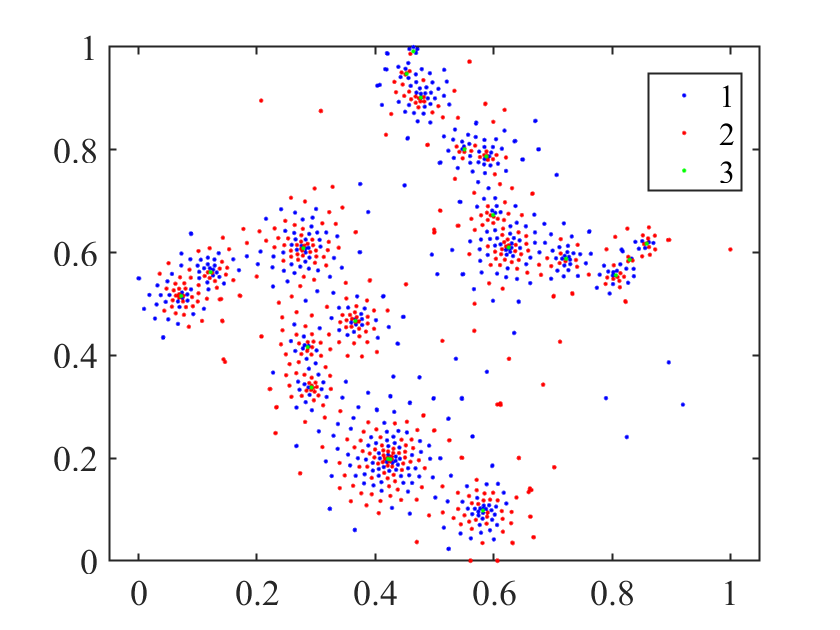}&
       % 0.56119      0.98005       2548.8
      \includegraphics[width=1.6in]{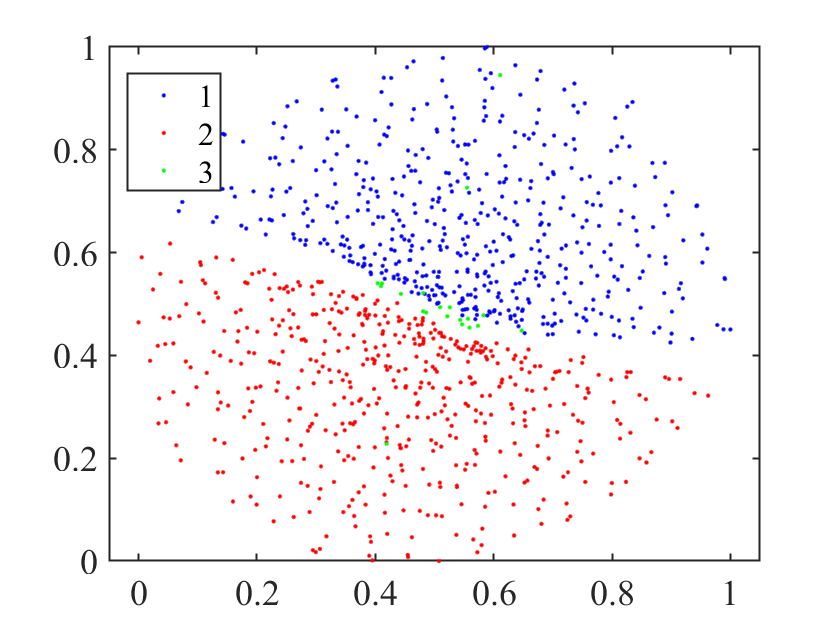} &
       % 0.17793       8.5203         1948
       \includegraphics[width=1.6in]{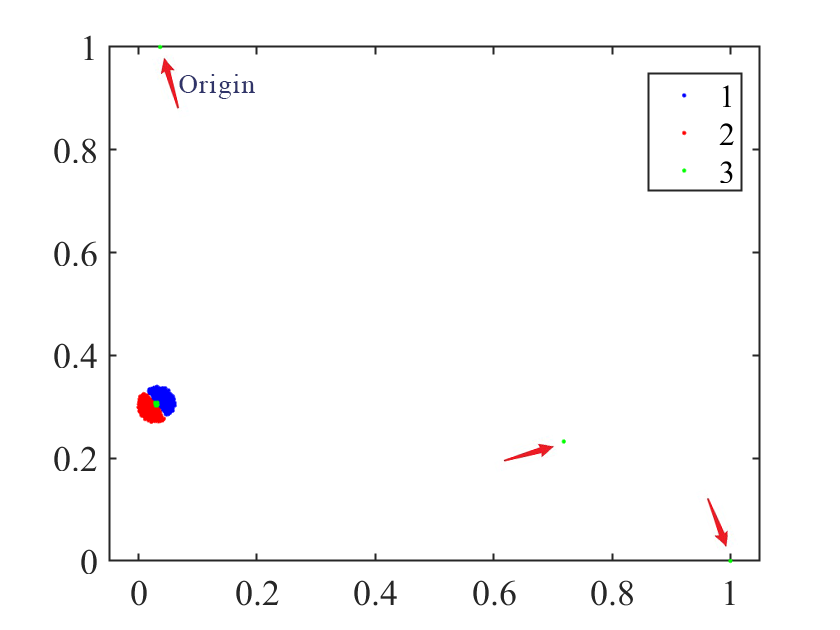}  \\     
      %      0.088882       7.7477       1923.2
      & (a) $perplexity=2$   & (b) $perplexity=50$ & (c) $perplexity=250$\\  \hdashline
       \begin{turn}{90} \quad  Isolation kernel \end{turn}&      \includegraphics[width=1.6in]{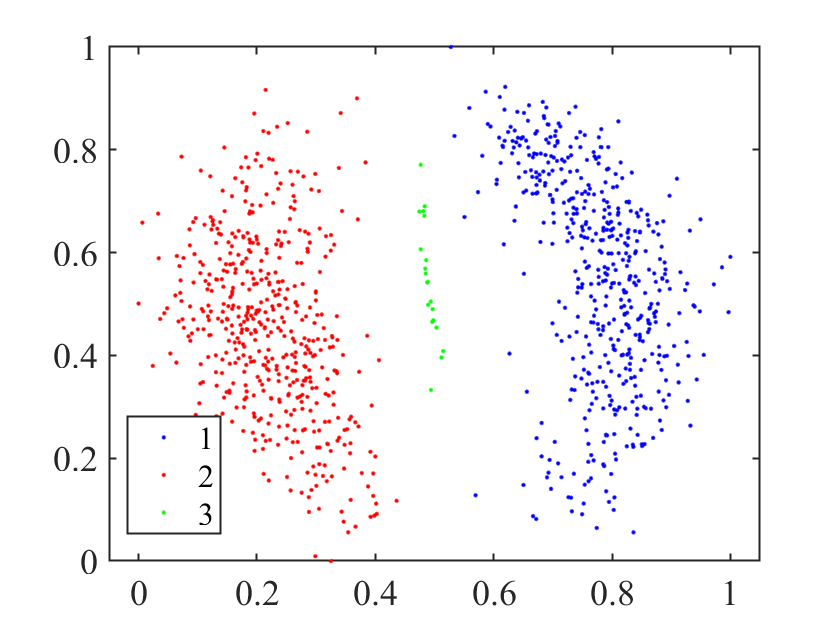} &
       %   0.063807      0.62081        16682
       \includegraphics[width=1.6in]{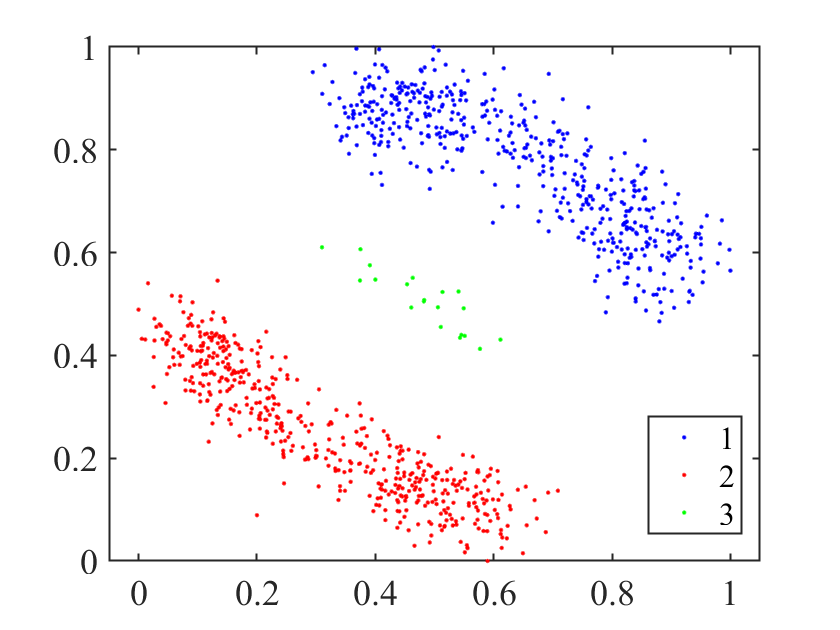} &
       % 0.21948      0.46932        10986
       \includegraphics[width=1.6in]{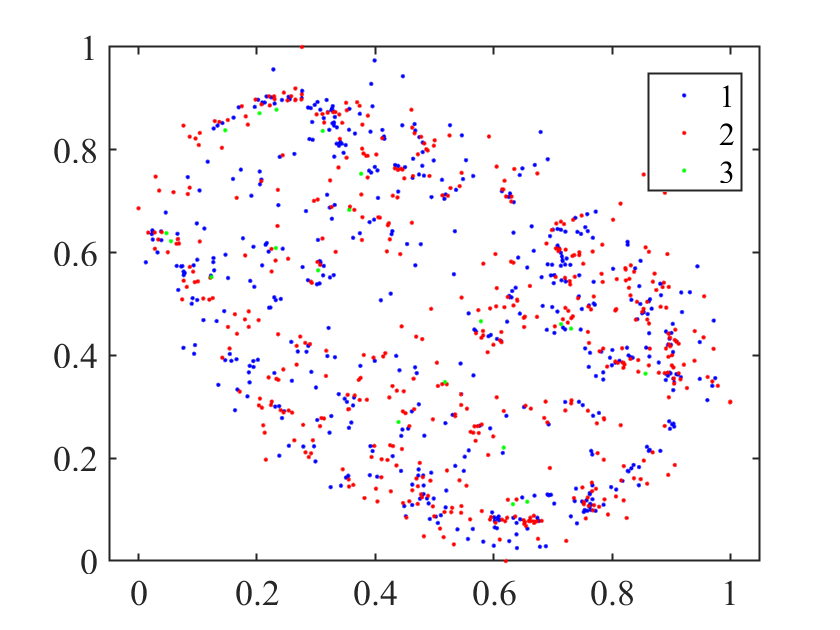}  \\    
       & (d) $\psi=2$ &  (e) $\psi=20$ & (f) $\psi=250$ \\
       %  0.52777      0.49528       4644.6
       \hline 
   \end{tabular}
 \label{sub1}
 \end{table}

\section*{Appendix B. The nearest neighbour implementation of the Isolation kernel} \label{app2}

We use an existing nearest neighbour method to implement the Isolation kernel \cite{aNNE}. It produces each partitioning $H$ (a Voronoi diagram) which consists of $\psi$ isolating partitions $\theta$, given a subsample $\mathcal{D}$ of $\psi \ge 2$ points.  Each isolating partition or Voronoi cell $\theta \in H$ isolates one data point from the rest of the points in the subsample. The point which determines a cell is called the cell centre. The Voronoi cell centred at $z \in \mathcal{D}$ is given as:
\[ \theta[z] = \{x  \in \mathbb{R}^d \ | \  z = \argmin_{{z} \in \mathcal{D}} \ell_p(x - {z})\}. \]
\noindent where $\ell_p(x, y)$ is a distance function and we use $p=2$ as Euclidean distance in this paper. 
 
\begin{table} [!htbp] 
\centering
  \begin{tabular}{cccc}
    \hline
     &  Uniform density distribution   & Parkinson dataset    \\
      \hline
 %     \hdashline 
       \begin{turn}{90}  \qquad  $\psi=16$  \end{turn}&       
       \includegraphics[width=1.8in]{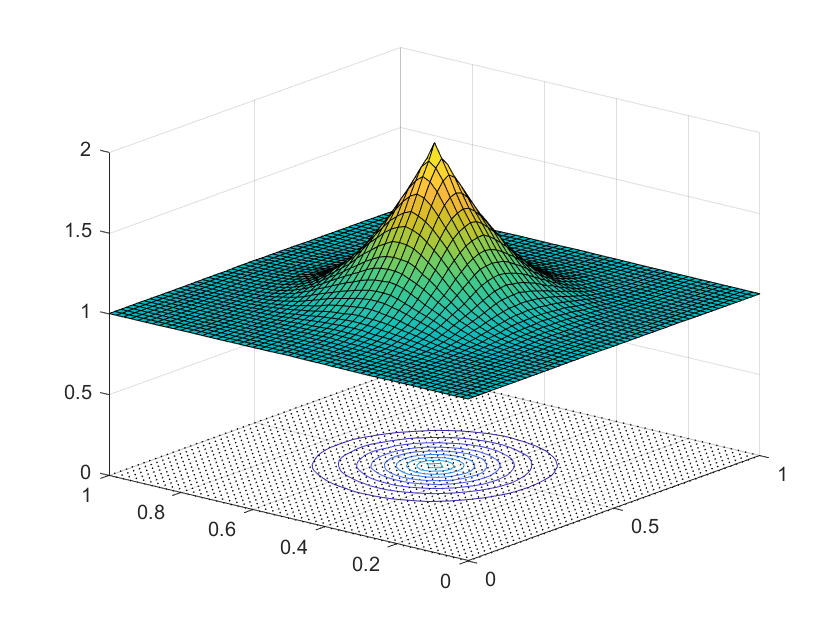}            &
      \includegraphics[width=1.8in]{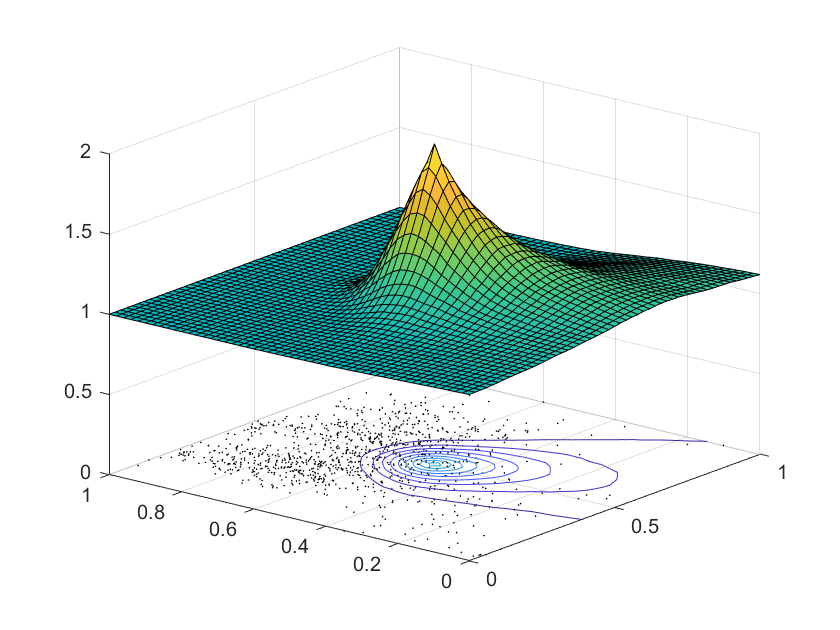}  \\   
    \hdashline 
       \begin{turn}{90} \qquad   $\psi=64$  \end{turn}&       
          \includegraphics[width=1.8in]{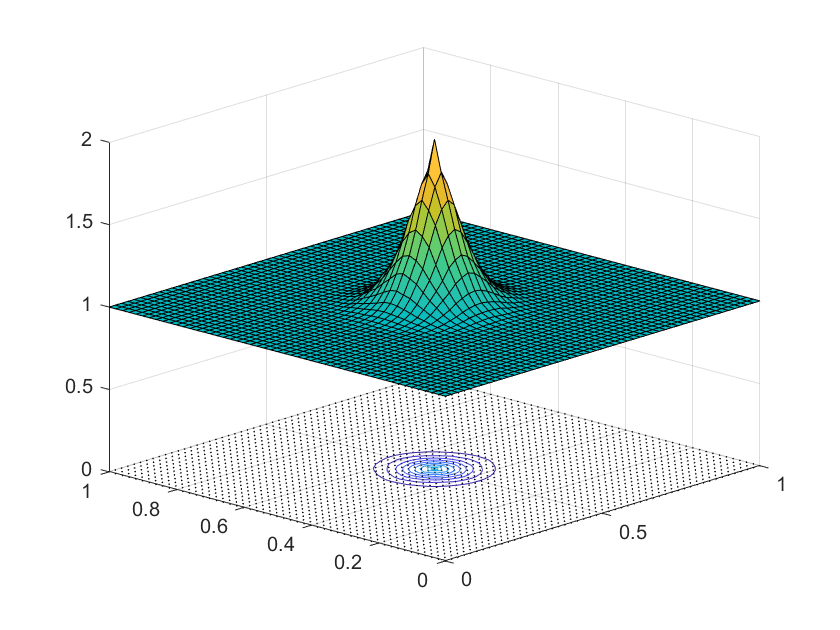}                           &
      \includegraphics[width=1.8in]{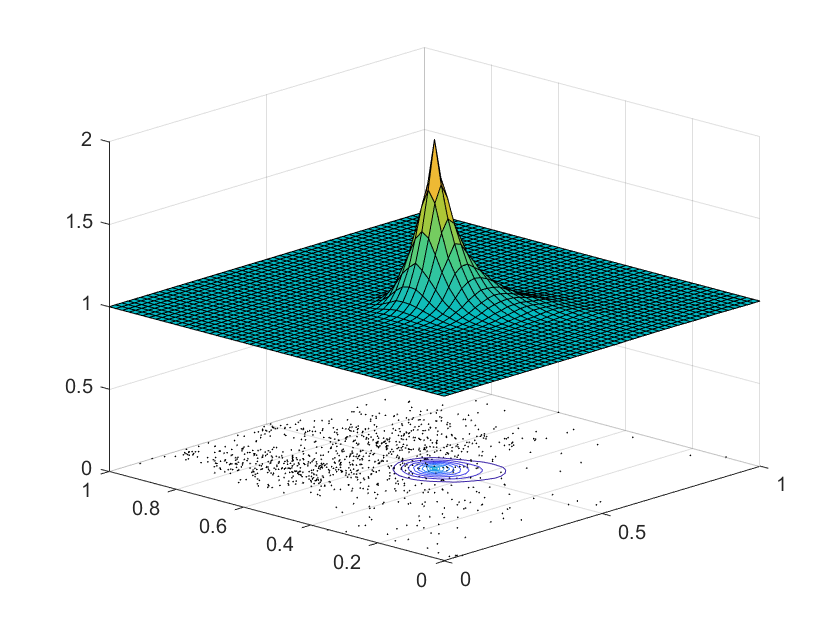}  \\     
    \hdashline 
       \begin{turn}{90} \qquad    $\psi=128$  \end{turn}&       
          \includegraphics[width=1.8in]{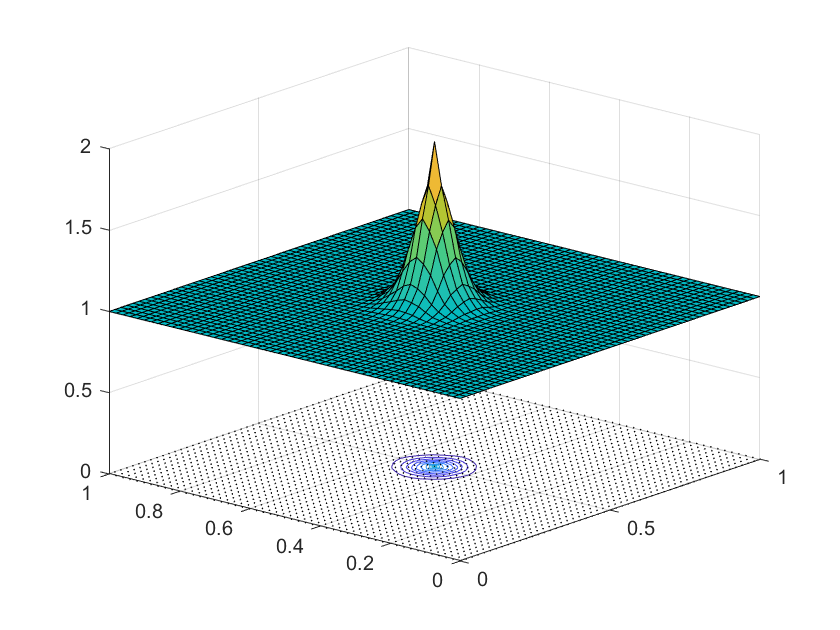}                           &
      \includegraphics[width=1.8in]{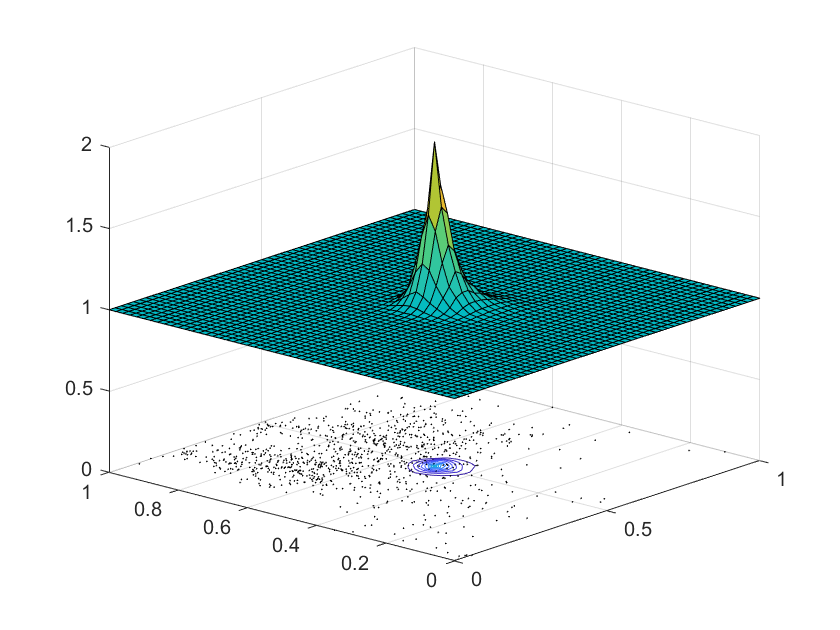}  \\    
      \hline
  \end{tabular}
  \caption{Contours of the Isolation kernel with reference to point (0.5, 0.5) on 2-dimensional datasets for three different values of $\psi$. Parkinson dataset uses 12th vs 21st attributes.}
  \label{illus}
\end{table}

Table \ref{illus} compares the contours of the Isolation kernel on two different data distributions with different $\psi$ values. It shows that the Isolation kernel is adaptive to the local density. Under uniform data distribution, the Isolation kernel's contour is symmetric with respect to the reference point at (0.5, 0.5). However, on the Parkinson dataset, the contour shows that, for points having equal inter-point distance from the reference point $x$ at (0.5, 0.5), points in the spare region are more similar to $x$  than points in the dense region to $x$. In addition, the larger the $\psi$, the sharper the kernel distribution of the Isolation kernel, as shown in Table \ref{illus}. This is because a larger $\psi$ produces more partitions (or Voronoi cells) of smaller sizes. This means that two points are less likely to fall into the same cell unless they are very close.

While this implementation of the Isolation kernel produces its contour similar to that of an exponential kernel  $k(x,y) = exp(-\frac{\parallel x-y \parallel}{2\sigma^2})$ under uniform density distribution, different implementations have different contours. For example, using axis-parallel partitionings to implement the Isolation kernel produces a contour (with the diamond shape) which is more akin to that of Laplacian kernel $k(x,y) = exp(-\frac{\parallel x-y \parallel}{\sigma})$ under uniform density distribution \cite{ting2018IsolationKernel}. Of course, both the exponential and Laplacian kernels, like Gaussian kernel, are data-independent.

\section*{Appendix C. t-SNE visualisation on News20} \label{app}

We compare the visualisation results of News20 with different parameter settings in Table \ref{v5}. It is interesting to note that t-SNE using the Isolation Kernel having a small $\psi$ produces better visualisation results having more separable clusters than those using Gaussian kernel with high perplexity, although the IK got slightly lower evaluation measure values (compare Figures (c) and (d) in each of Tables \ref{v5}.) However, the two clusters are significantly overlapped in most cases.

\begin{table} [!htb]  
\centering
\caption{Visualisation result of t-SNE on News20. t-SNE produced the best DB scores when using Gaussian Kernel with $perplexity=3700$ and  Isolation Kernel with $\psi=85$.} 
  \begin{tabular}{|c|ccc|} 
      \hline
 %     \hdashline
       \begin{turn}{90} \  Gaussian Kernel \end{turn}&      \includegraphics[width=1.7in]{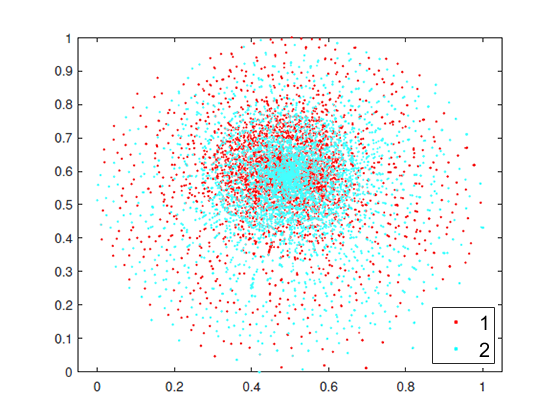}&
      \includegraphics[width=1.7in]{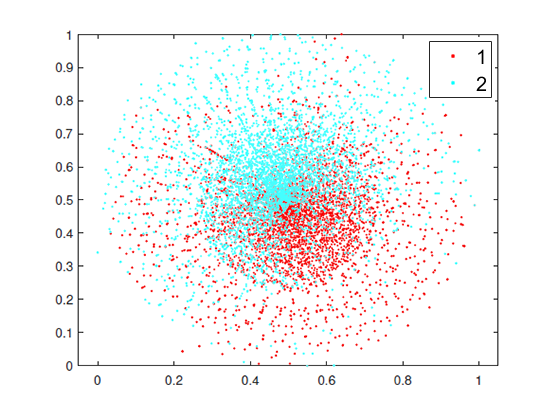} &
      \includegraphics[width=1.7in]{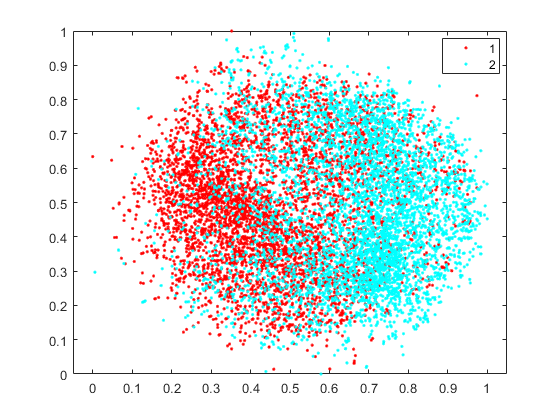}  \\     
       & (a) $perplexity=31$   & (b) $perplexity=910$ & (c) $perplexity=3700$\\  \hdashline
       \begin{turn}{90} \ \  Isolation Kernel \end{turn}&      \includegraphics[width=1.7in]{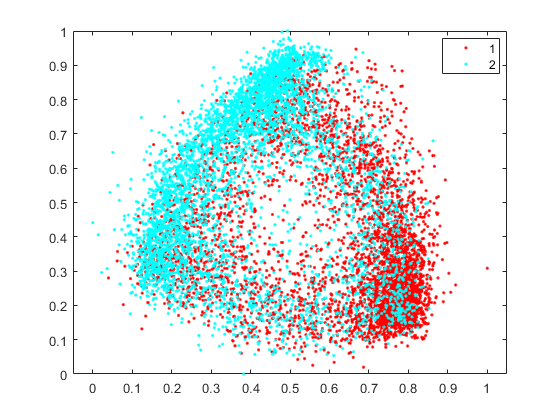} &
      \includegraphics[width=1.7in]{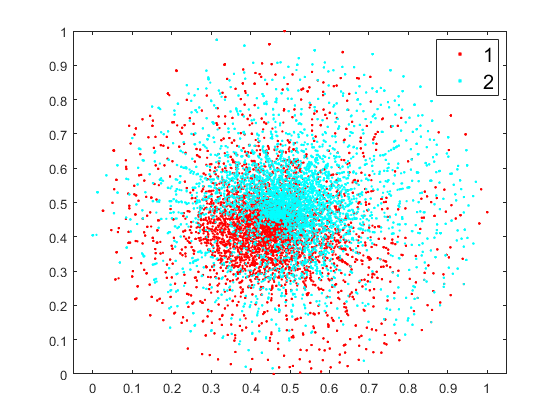} &
      \includegraphics[width=1.7in]{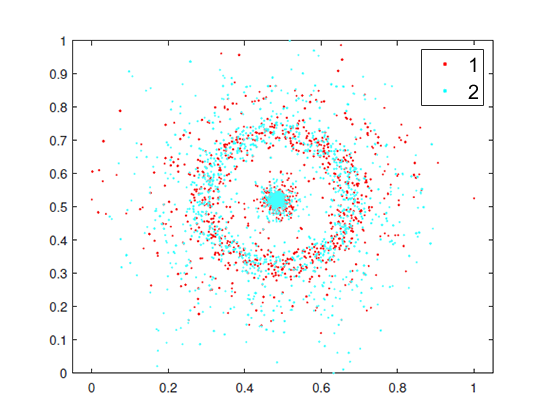}  \\    
      & (d) $\psi=85$ &  (e) $\psi=128$ & (f) $\psi=3000$ \\
      \hline 
  \end{tabular}
\label{v5}
\end{table}

\begin{figure}[!hbt]
\centering
    \includegraphics[width=3in]{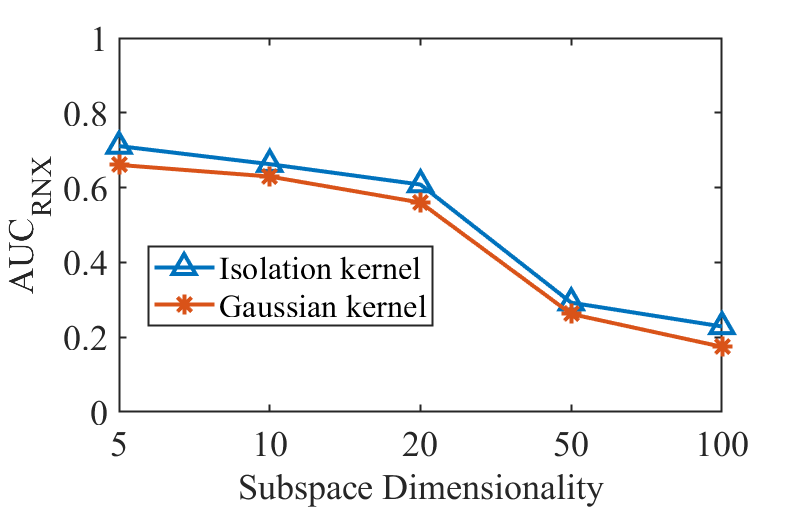}
  \caption{$AUC_{RNX}$ of Gaussian Kernel and Isolation Kernel on 5 subspace clusters with different dimensionality. The parameters for each algorithm are turned according to Table~\ref{para}.}
   \label{sub}  
\end{figure}

We suspect that the overlapping issue is caused by the sparsity. To verify it, we use the same data distribution from Table \ref{v0} and increase the dimensionality of the 5 subspace clusters. The results in Table \ref{sub} show that t-SNE with both kernel measures got lower $AUC_{RNX}$ scores when increasing the cluster dimensionality, i.e,. clusters are becoming more sparse in higher dimensional space.

\section*{Appendix D. Other data-dependent kernels} \label{app0}

Recall that the first step of t-SNE may be interpreted as using kNN to determine the $n$ bandwidths of Gaussian kernel. There are existing kNN-based data-dependent kernels that adapt to local density, i.e.,

\renewcommand{\labelenumi}{\roman{enumi})}
\begin{enumerate}
 \item \textbf{kNN kernel} \cite{KNNKernel}.
 
 The kNN kernel is a binary function defined as:
  \begin{eqnarray}
K_{kNN}({ x},{ y})  & = &     \mathds{1}( y \in kNN( x))  
 \label{Eqn_kNN}
\end{eqnarray}
 \noindent where $kNN(x)$ is the set of $k$ nearest neighbours of $x$.

 \item \textbf{Adaptive Gaussian kernel} \cite{zelnik2005self}. 
 
 The distance of $k$-th NN has been used to set the bandwidth of Gaussian kernel to make it adaptive to local density. This was proposed in spectral clustering as an adaptive means to adjust the similarity to perform dimensionality reduction before clustering. 
 
 Adaptive Gaussian kernel is defined as:
   \begin{eqnarray}
K_{AG}({ x},{ y})  & = &   exp\frac{-|| x- y||^2}{\sigma_{ x} \sigma_{ y}}
 \label{Eqn_AG}
\end{eqnarray}
\noindent where $\sigma_{ x}$ is the distance between $ x$ and $ x$'s $k$-th nearest neighbour.
\end{enumerate}

However, replacing the Gaussian kernel in t-SNE with either of these kernels produces poor outcomes. For example, on the Segment and Spam datasets, the adaptive Gaussian kernel produced AUC$_{RNX}$ scores of 0.35 and 0.22, respectively; and the kNN kernel yielded AUC$_{RNX}$ scores of 0.38 and 0.28, respectively. They are significantly poorer than those produced using the Gaussian kernel or Isolation kernel shown in Table \ref{tbl_results}. We postulate that this is because a global $k$ is unable to make these kernels sufficiently adaptive to local distribution.  

It is interesting to note that the current method used to get a data-dependent kernel is to begin with a data-independent kernel such as Gaussian kernel. And then find ways to make the Gaussian kernel data-dependent. This is an indirect approach.
%which often requires an optimisation, and sometimes not defined for any points in $\mathbb{R}^d$ (as in the case of the similarity used in t-SNE.) As a result, the data-dependent characteristic is often not well-defined. 
The Isolation kernel is a direct approach in getting a data-dependent kernel, derived directly from a given dataset, without an intermediary of a data-independent kernel.  
%Its advantage is clear: the data-dependent characteristic is well-defined, without the need of an optimisation. 

\section*{Appendix E. Visualisation results of Fast interpolation-based t-SNE}  \label{app3}

FIt-SNE \cite{linderman2019fast} addresses the runtime issue in step 3 of the t-SNE algorithm only.

\begin{figure}[!hbt]
\centering
  \begin{subfigure}{.49\textwidth}
  \centering
    \includegraphics[width=2in]{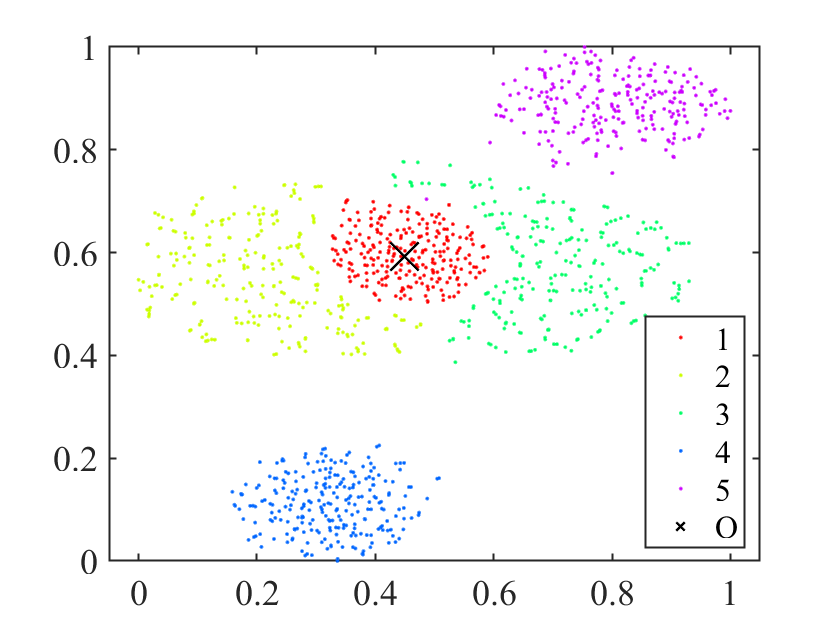}
  \caption{5 subspace clusters connected at one point}
  \end{subfigure}  %
  \begin{subfigure}{.49\textwidth}
  \centering
    \includegraphics[width=2in]{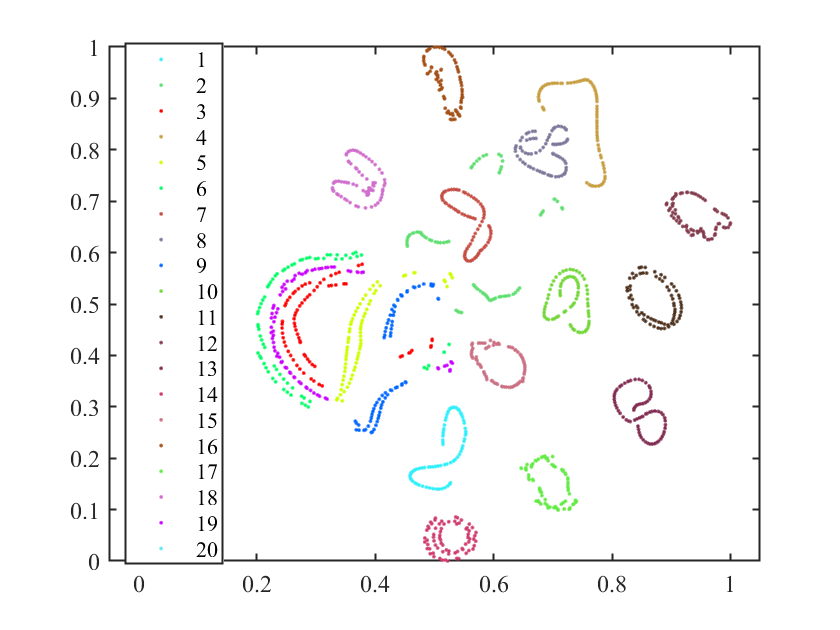}
  \caption{COIL20}
  \end{subfigure}
\caption{Visualisation of FIt-SNE on two datasets.}
    \label{ill2}  
\end{figure}

Figure \ref{ill2} demonstrates the visualisation results of FIt-SNE \cite{linderman2019fast} on two datasets. It is clear that FIt-SNE has the same deficiency of misrepresented structures as in t-SNE, due to the use of Gaussian kernel, as discussed in Section \ref{sec_pint-based_Bandwidth}.

\begin{figure}[!hbt]
\centering
  \begin{subfigure}{.43\textwidth}
  \centering
    \includegraphics[width=2in]{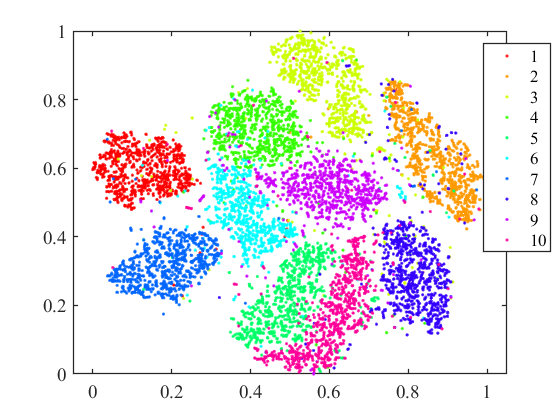}
  \caption{MNIST in all steps, CH=5926}
  \end{subfigure}  \\
  \begin{subfigure}{.43\textwidth}
  \centering
    \includegraphics[width=2in]{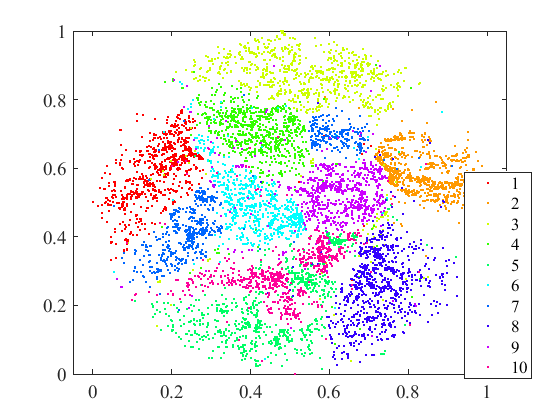}
  \caption{MNIST8M (step 1) \& MNIST (steps 2, 3), CH=3529}
  \end{subfigure} \hspace{5mm} 
    \begin{subfigure}{.43\textwidth} 
  \centering
    \includegraphics[width=2in]{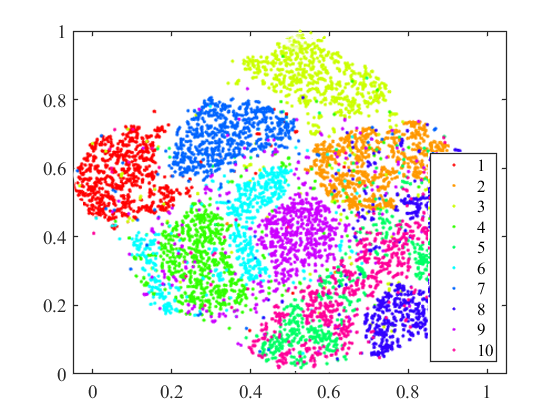}
  \caption{MNIST8M(step 1) \& 10,000 (steps 2, 3), CH=2973}
  \end{subfigure}
  \caption{FIt-SNE visualisation results with the Gaussian kernel on the MNIST and MNIST8M datasets. }
 \label{scale2}  
\end{figure}

Figure \ref{scale2} shows the FIt-SNE results on the MNIST and MNIST8M datasets.\footnote{The source code is obtained from \url{https://github.com/KlugerLab/FIt-SNE}. All parameters in FIt-SNE are set to the default except that we search for the best $perplexity$ in the same range as t-SNE in Table~\ref{para}.}  FIt-SNE's results are worse than those of  t-SNE based on either GK or IK in terms of the CH scores on both the MNIST and MNIST8M datasets; so as the visualisation outcomes. Note that without the colours to differentiate between classes, most of the classes shown in Figure \ref{scale2} cannot be identified as separate classes in the FIt-SNE results produced from the MNIST8M dataset.

FIt-SNE ran faster than t-SNE because of approximation using grid;
%with only 17s using  
and it is implemented with C++ with multi-threading. The price it paid to be more efficient using the approximation is worse visualisation outcomes.

Note that on the MNIST8M dataset, we could only use 2 million data points in FIt-SNE because of its high memory usage. In contrast, with the Isolation kernel, we could run t-SNE (in MatLab without multithreading) on the same machine using the entire 8.1 million data points of MNIST8M (shown in Table \ref{scale}.)

\section*{Appendix F. Visualisation results of opt-SNE}  
opt-SNE \cite{belkina2019automated} is an enhanced version of t-SNE that aims to improve the local structure resolution and produce a reliable embedding of a dataset.  However, since opt-SNE still uses Gaussian kernel, it has the same deficiency of misrepresented structures as the original t-SNE.

\begin{figure}[!hbt]
\centering
  \begin{subfigure}{.5\textwidth}
  \centering
    \includegraphics[width=2in]{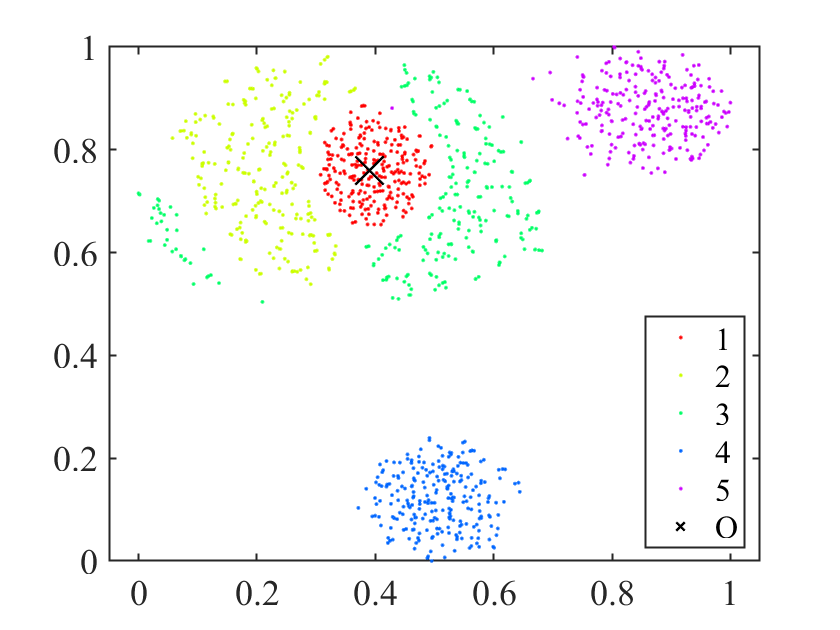}
  \caption{5 subspace clusters connected at one point}
   \label{scale3:a}  
  \end{subfigure}  \\
  \begin{subfigure}{.45\textwidth}
  \centering
    \includegraphics[width=2in]{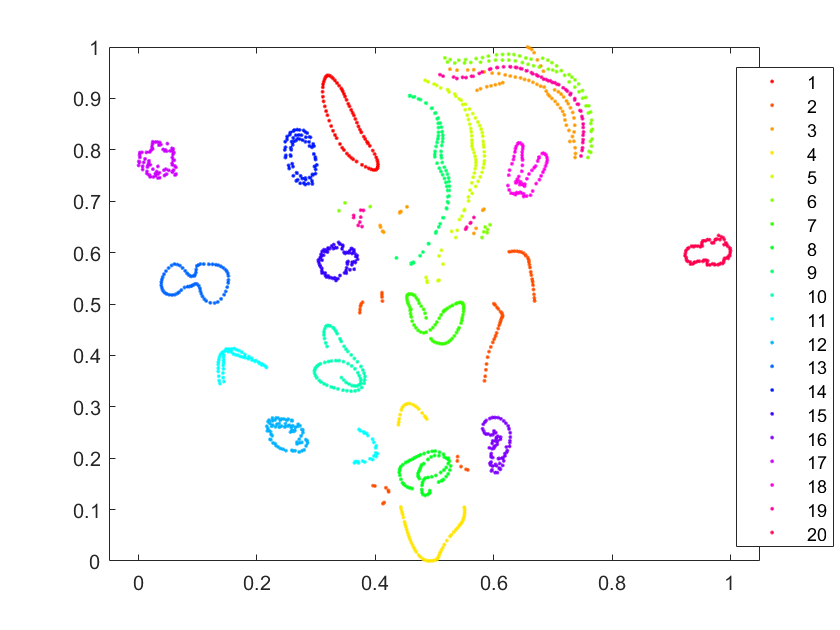}
  \caption{COIL20}
  \label{scale3:b} 
  \end{subfigure} \hspace{5mm} 
    \begin{subfigure}{.45\textwidth} 
  \centering
    \includegraphics[width=2in]{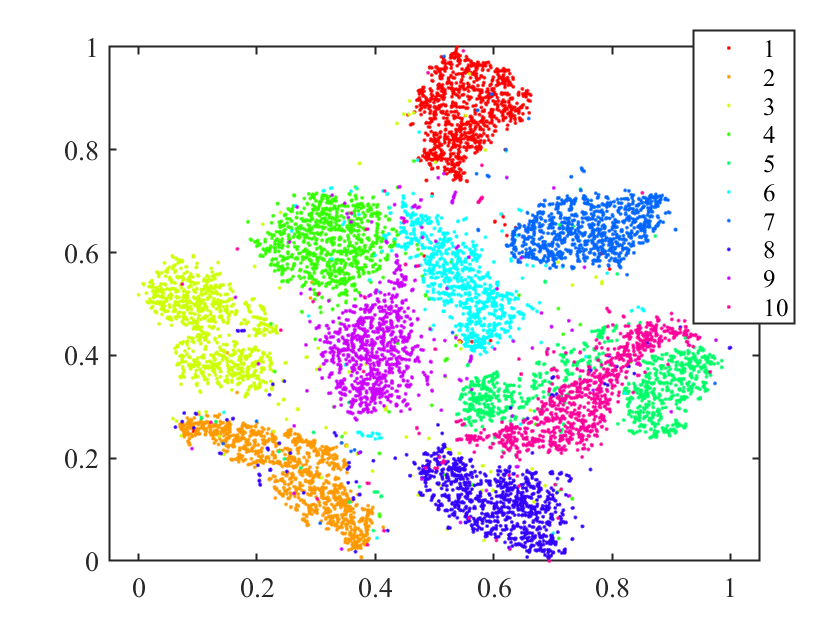}
  \caption{MNIST} % CH= 6129
  \label{scale3:c}
  \end{subfigure}
  \caption{opt-SNE visualisation results with Gaussian kernel on three datasets. }
 \label{scale3}  
\end{figure}

Figure \ref{scale2} shows the visualisation results on three datasets using opt-SNE \footnote{The source code is obtained from \url{https://github.com/omiq-ai/Multicore-opt-SNE}. All parameters in opt-SNE use the default settings except that we search for the best $perplexity$ in the same range as t-SNE stated in Table \ref{para}.}.    As expected, opt-SNE produced similar results as t-SNE, having misrepresented structures in Figures \ref{scale3:a} and  \ref{scale3:b}. On MNIST, opt-SNE got a slightly worse result than t-SNE  (CH=$6129$ versus CH=$6452$) because it split the green clusters into two parts, as shown in Figure~\ref{scale3:c}. 

\bibliographystyle{abbrv} 
\bibliography{ref}

\end{document}